\documentclass[sigconf]{acmart}
\AtBeginDocument{%
  }

\usepackage{enumitem}
\usepackage{algorithm}
\usepackage{algorithmic}

\usepackage{hyperref}       
\usepackage{url}            
\usepackage{booktabs}       
\usepackage{amsfonts}       
\usepackage{nicefrac}       
\usepackage{xcolor}         
\usepackage{amsmath}
\usepackage{makecell} 

\usepackage{multirow}
\usepackage{soul}
\usepackage{subcaption}
\usepackage{bm}
\usepackage{wasysym}

\begin{document}

\title{Evolving Demonstration Optimization for Chain-of-Thought Feature Transformation}

\author{Xinyuan Wang}
\affiliation{%
  \institution{Arizona State University}
  \city{Tempe}
  \state{Arizona}
  \country{USA}}
\email{xwang735@asu.edu}

\author{Kunpeng Liu}
\affiliation{%
  \institution{Clemson University}
  \city{Clemson}
  \state{South Carolina}
  \country{USA}}
\email{kunpenl@clemson.edu}

\author{Arun Vignesh Malarkkan}
\affiliation{%
  \institution{Arizona State University}
  \city{Tempe}
  \state{Arizona}
  \country{USA}}
\email{amalarkk@asu.edu}

\author{Yanjie Fu}
\affiliation{%
  \institution{Arizona State University}
  \city{Tempe}
  \state{Arizona}
  \country{USA}}
\email{yanjie.fu@asu.edu}
\renewcommand{\shortauthors}{Trovato et al.}

\begin{abstract}
Feature Transformation (FT) is a core data-centric AI task that improves feature space quality to advance downstream predictive performance. 
However, discovering effective transformations remains challenging due to the large space of feature-operator combinations.
Existing solutions rely on discrete \emph{search} or latent \emph{generation}, but they are frequently limited by sample inefficiency, invalid candidates, and redundant generations with limited coverage.
Large Language Models (LLMs) offer strong priors for producing valid transformations, but current LLM-based FT methods typically rely on \emph{static} demonstrations, resulting in limited diversity, redundant outputs, and weak alignment with downstream objectives.
We propose a framework that optimizes \emph{context data} for LLM-driven FT by evolving trajectory-level experiences in a closed loop.
Starting from high-performing feature transportation sequences explored by reinforcement learning, we construct and continuously update an experience library of downstream task-verified transformation trajectories, and use a diversity-aware selector to form contexts along with a chain-of-thought and guide transformed feature generation toward higher performance.
Experiments on diverse tabular benchmarks show that our method outperforms classical and LLM-based baselines and is more stable than one-shot generation. 
The framework generalizes across API-based and open-source LLMs and remains robust across downstream evaluators.
Codes are \href{https://anonymous.4open.science/r/CoT_Feature_Transformation-40B4/}{provided}.

\end{abstract}

\renewcommand\footnotetextcopyrightpermission[1]{} 
\settopmatter{printacmref=false}


\maketitle

\section{Introduction}

Data-centric AI aims to improve the quality of data to advance the performance and generalization of ML.
As an essential tool, Feature Transformation (FT) derives new features by combining original features with operators (e.g., transforming $[a,b]$ into $[a/b,\; a-b,\; (a+b)/a]$).
FT can reconstruct distance measures, reshape discriminative patterns, and enhance data AI readiness (e.g., structural, predictive, interaction, and expression levels). 

\begin{figure}[htbp]
    \centering
    \begin{subfigure}[b]{0.48\linewidth}
        \includegraphics[width=\linewidth]{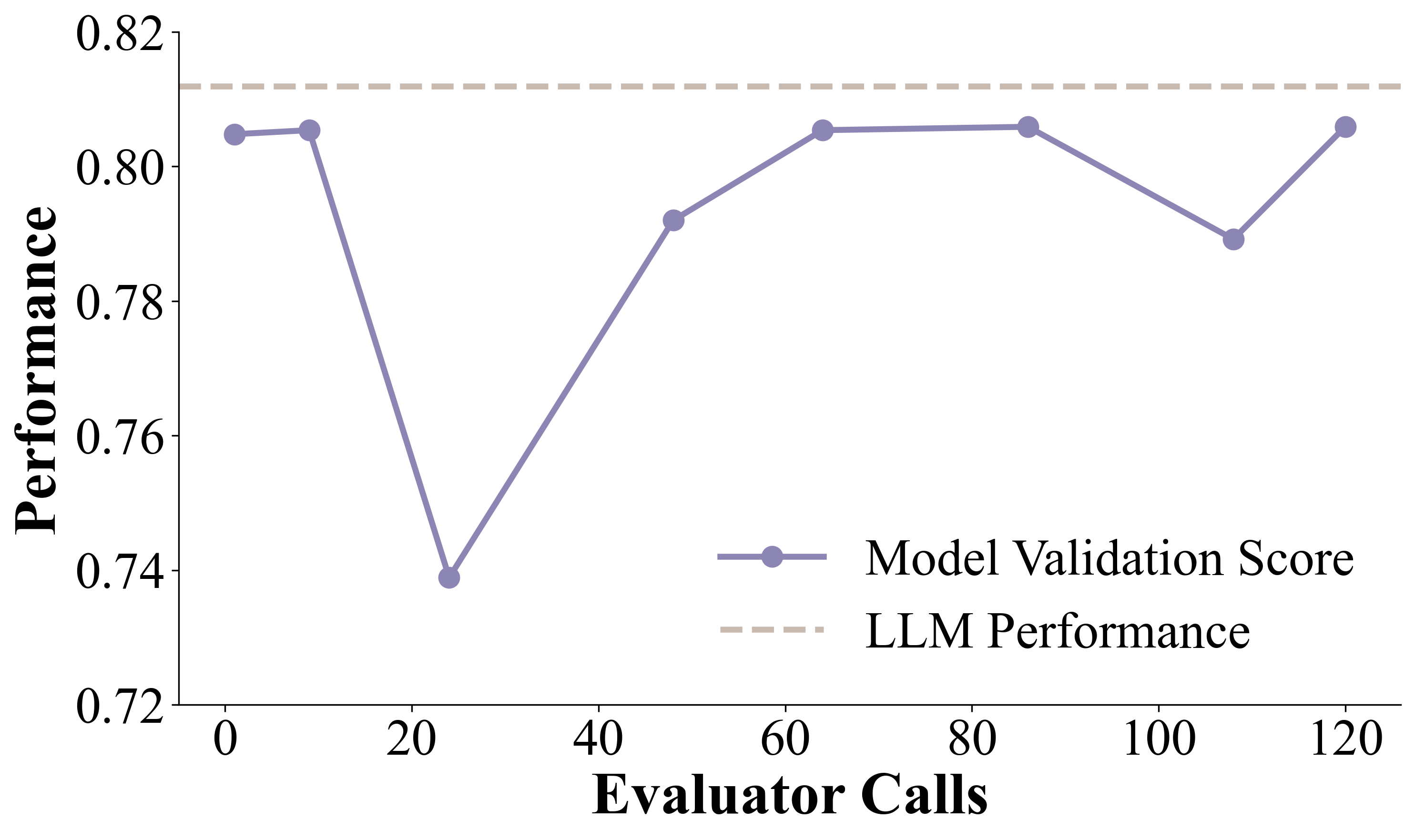}
        \caption{Calls of Search.}
        \label{fig:motivation_ml_search}
    \end{subfigure}
    \hfill
    \begin{subfigure}[b]{0.48\linewidth}
        \includegraphics[width=\linewidth]{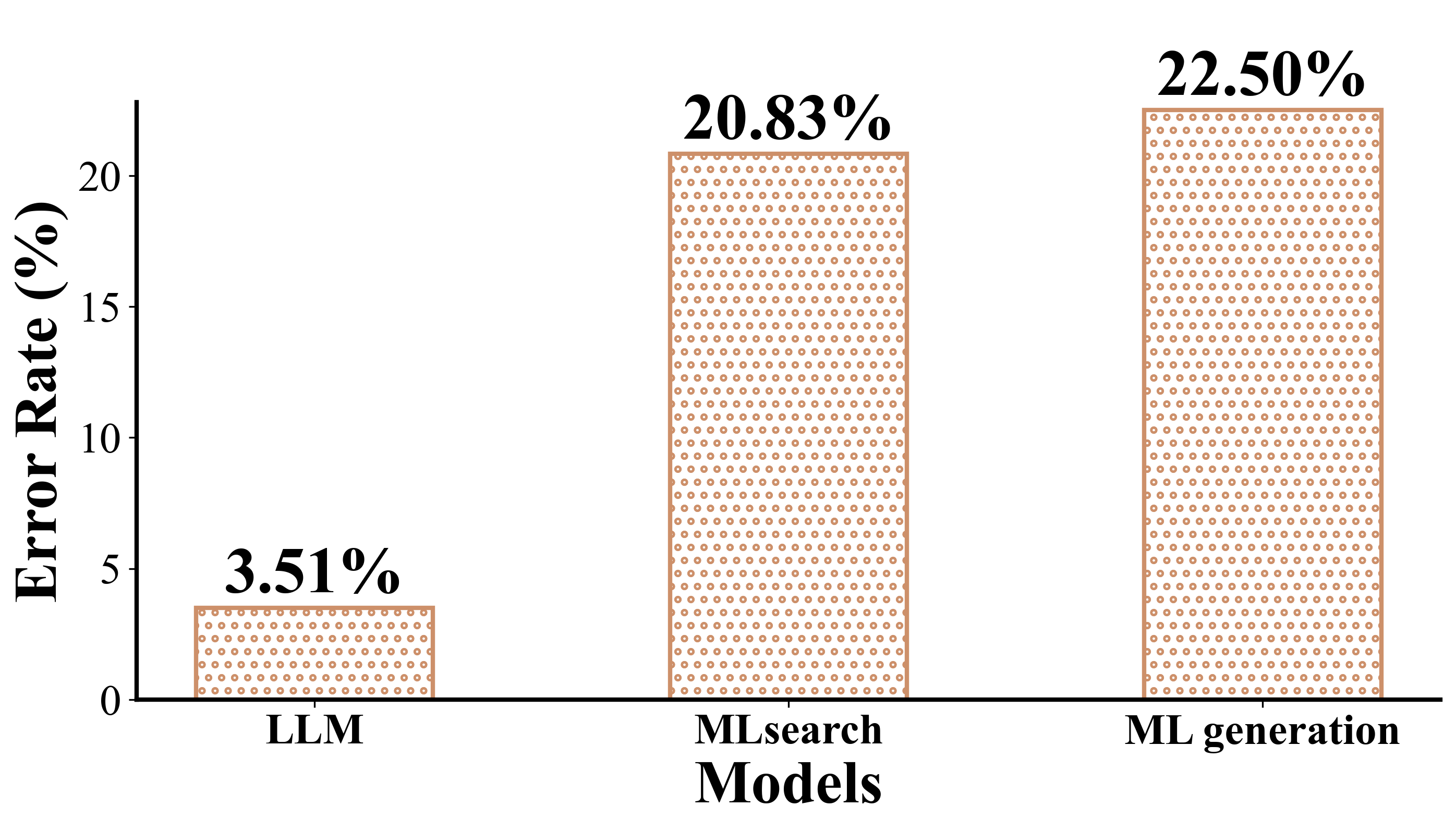}
        \caption{Invalid Ratio.}
        \label{fig:motivation_ml_generation}
    \end{subfigure}
    \hfill
    \begin{subfigure}[b]{0.8\linewidth}
        \includegraphics[width=\linewidth]{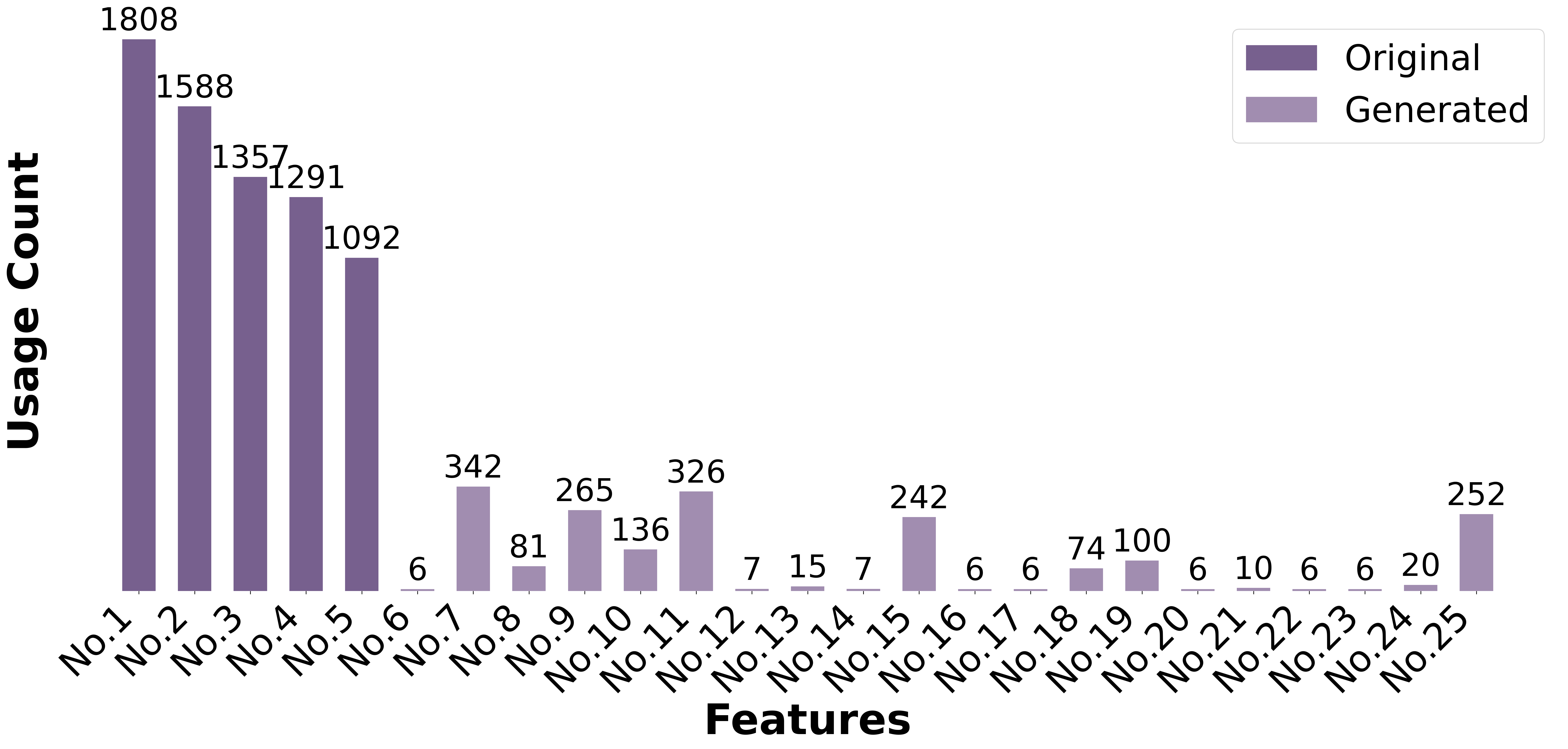}
        \caption{Redundancy of Fixed Prompt.}
        \label{fig:motivation_llm_stable}
    \end{subfigure}
    \caption{Empirical motivations.}
    \label{fig:motivation}
\end{figure}

Existing FT methods follow the traditional machine learning paradigm.
i) Discrete search-based methods (e.g., GRFG~\cite{wang2022group}) employ reinforcement learning agents to make discrete actions on features and operators, leading to new feature combinations. However, the space of compositional transformations grows combinatorially with feature dimension and operator choices, making exhaustive search infeasible. And agents often suffer from sparse rewards and biased exploration (\textbf{Figure~\ref{fig:motivation_ml_search}} shows too many calls and unstable performance of the search methods).
ii) Latent space-based generative methods (e.g., MOAT~\cite{wang2023reinforcement}), based on the RL exploration, optimize continuous embeddings and search for better ones to decode new feature combinations. 
But both methods have weak dataset background knowledge, so the ``blind'' search may cause a mismatch between training and search, and it often fails to produce transformations that are both valid and executable. 
For example, \textbf{Figure~\ref{fig:motivation_ml_generation}} shows a higher invalid ratio of existing methods.

Recent progress in generative AI suggests that Large Language Models (LLMs) can produce richer, more diverse feature combinations than manual design or heuristic search~\cite{zhao2023survey, xie2025text}.
The view of generative feature transform (GFT)~\cite{wang2025towards,ying2025survey} treats a newly transformed feature as a \emph{feature combination} (e.g., $a/b$), a transformed feature set as a \emph{feature combination token sequence}, and the creation of a new feature set as a token-sequence generation problem.
This perspective replaces exponential search with conditional generation, and the knowledge of LLMs can prevent invalid combinations.
Existing LLM-driven prompt-based methods~\cite{hollmann2023large,gong2024evolutionary} generate transformations through in-context prompting equipped with fixed few-shot experience. 
However, most methods rely on static prompting: few-show examples are fixed (or lightly edited) and are not improved by downstream feedback. The useful transformation patterns are not accumulated across steps.
In other words, \emph{LLMs provide priors}, but the prompts that inject these priors are still treated as static and non-optimized (\textbf{Figure~\ref{fig:motivation_llm_stable}} shows that LLMs prefer specific features, which limits the diversity and creativity).

\begin{figure}[htbp] 
    \centering 
    \includegraphics[width=1.0\linewidth]{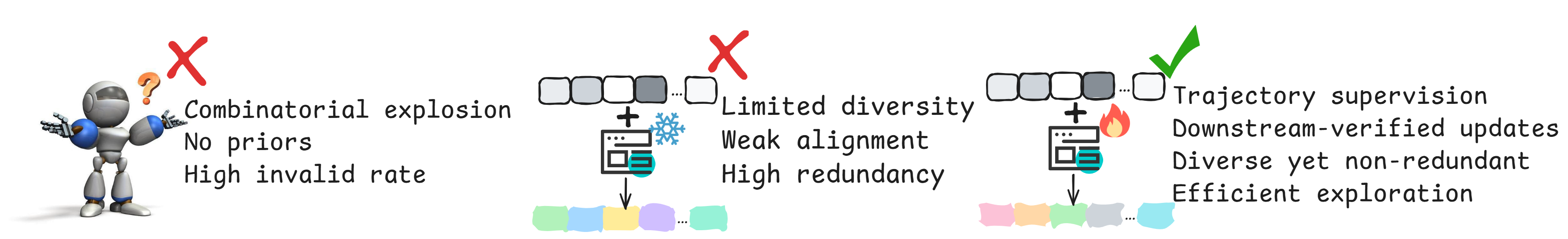} 
    \caption{Comparison of different solutions.} 
    \label{fig:motivation_comparsion} 
\end{figure}

For LLM-driven FT, the most controllable and reusable ``data'' is not only a tabular dataset. However, we consider the few-shot demonstrations that encode transformation experiences and are fed to the LLM as important.
Recent findings show that optimizing an evolving experiential context can prompt a frozen LLM by treating experience as a token prior (i.e., shifting optimization from parameter space to context space)~\cite{cai2025trainingfree}.
Motivated by this, we treat few-shot examples as an \emph{evolving experience library} rather than static prompts (\textbf{Figure~\ref{fig:motivation_comparsion}}).
Specifically, we build few-shot demonstrations through a closed-loop, data-centric pipeline: we first explore high-performing transformation sequences via reinforcement learning; then refine them at three levels, including (i) combination-wise validity checking, (ii) merging multiple examples into performance-sorted transformation trajectories in a chain-of-thought style with outlier filtering and LLM-based enhancement to guide the next iteration, and (iii) entropy-based diversity selection to ensure broad coverage with low redundancy.
This turns prompt construction into a principled data-centric optimization process aligned with downstream performance.

The results on diverse domains show improvements over classical search-based and automated feature engineering baselines, and stronger, more stable gains than one-shot LLM generation under the same evaluation budget, with ablations highlighting the importance of CoT-style refinement. We further demonstrate transferability across both API-based and open-source LLMs, robustness across downstream evaluators, and interpretable LLM behaviors that motivate our diversity control.

There are three key contributions for our method: \textbf{(1) Context-as-data formulation for GFT:} We see LLM-driven feature transformation as a data-centric optimization problem where few-shot demonstrations function as reusable, adaptable experiences rather than fixed prompts. \textbf{(2) Closed-loop experience construction:} We propose a loop that explores, evaluates, refines, and constructs transformation trajectories into an evolving demonstration library, enabling non-invasive LLM performance. \textbf{(3) View of dynamic transformation trajectory:} We treat the signal examples as parts of a trajectory rather than a stable experience, which shows the evolution path aligned with downstream performance to LLM.

\section{Preliminaries and Problem Statement}

\subsection{Important Concepts}
\label{sec:concept}




\noindent\textbf{Operation Set:} To refine the feature space, we need to apply mathematical operations to existing features to generate new informative features. All operations are collected in an operation set, denoted by $\mathcal{O}$. These operations can be classified as unary and binary operations. The unary operations such as ``\texttt{square}'', ``\texttt{exp}'', ``\texttt{log}'', etc. The binary operations such as ``\texttt{plus}'', ``\texttt{multiply}'', ``\texttt{minus}'', etc.

\begin{figure}[htbp] 
    \centering 
    \includegraphics[width=0.8\linewidth]{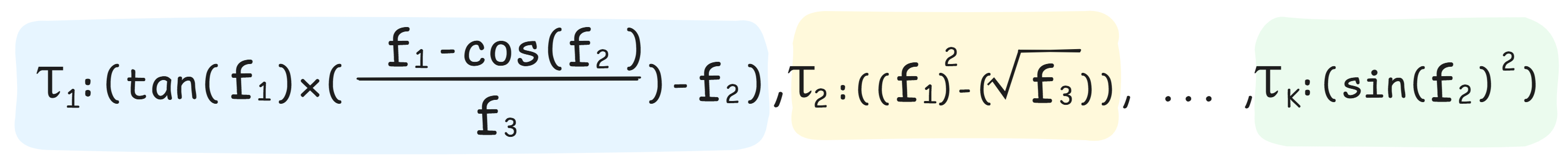} 
    \caption{A feature transformation sequence example.} 
    \label{fig:app_example_seq} 
\end{figure}

\noindent\textbf{Feature Transformation Sequence:} Assume that a dataset $\mathcal{D}=\{(\mathbf{x}^{(i)}, y^{(i)})\}_{i=1}^{N}$ includes the original feature set $\mathcal{F}_0=\{f_1,\ldots,f_K\}$ and predictive targets $y$. We transform existing features using mathematical combinations $\tau$ consisting of feature IDs and operations to generate new and informative features (\textbf{Figure~\ref{fig:app_example_seq}}). $K$ combinations are adopted to refine $\mathcal{F}_0$ to a better feature space $\tilde{\mathcal{F}_0}=\{\tilde{f}_{1},\ldots,\tilde{f_K}\}$. The collection of $K$ combinations refers to the feature transformation sequence, which is denoted by $\Gamma = [\tau_{1},\cdots,\tau_{K}]$.

\begin{figure}[htbp]
    \centering
    \captionsetup{justification=centering,singlelinecheck=false} 
    \begin{subfigure}[b]{0.45\linewidth}
        \includegraphics[width=\linewidth]{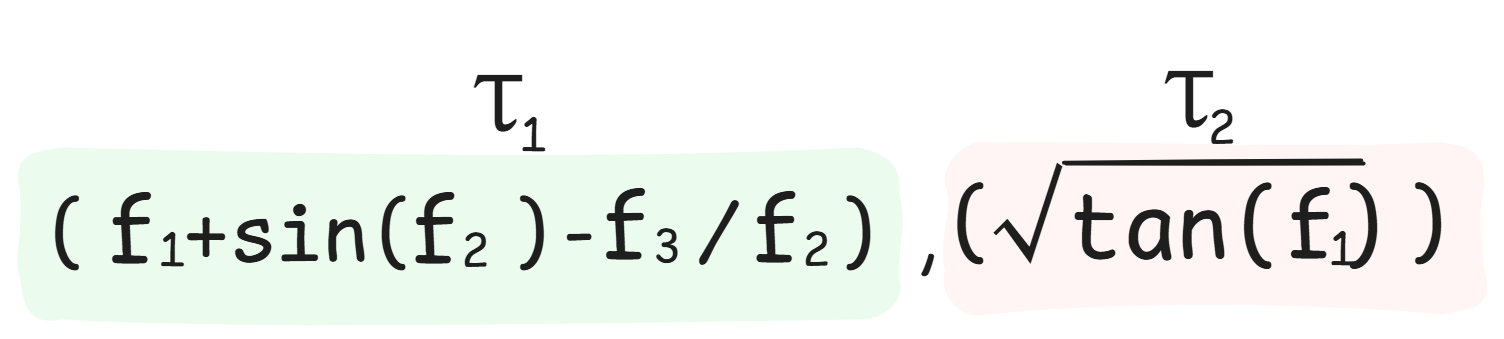}
        \caption{Original Sequence}
        \label{fig:example_postfix_a}
    \end{subfigure}
    \hfill
    \begin{subfigure}[b]{1.0\linewidth}
        \includegraphics[width=\linewidth]{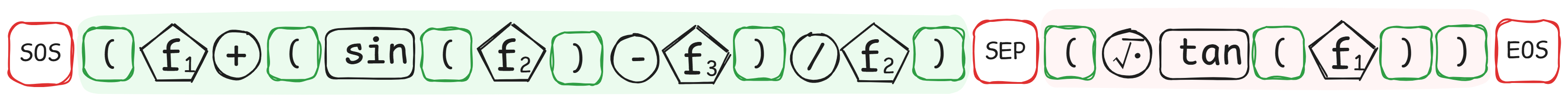}
        \caption{Infix Expression}
        \label{fig:example_postfix_b}
    \end{subfigure}
    \hfill
    \begin{subfigure}[b]{0.68\linewidth}
        \includegraphics[width=\linewidth]{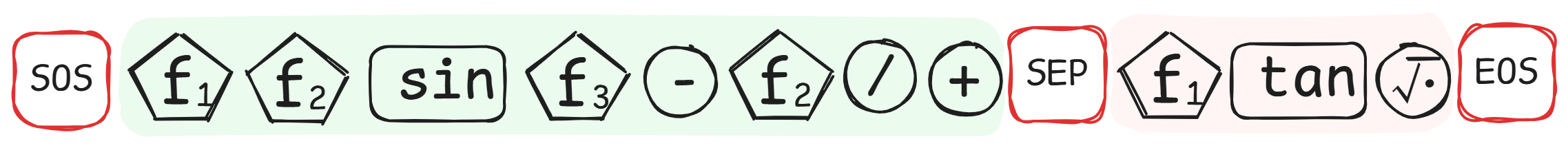}
        \caption{Postfix Expression}
        \label{fig:example_postfix_c}
    \end{subfigure}
    \caption{Different expressions of transformation sequences.}
    \label{fig:example_postfix_all}
\end{figure}

\noindent\textbf{Postfix Expressions:} 
The transformation sequence should be in a computable and machine-learnable format. \textbf{Figure~\ref{fig:example_postfix_a}} shows a transformation sequence with two generated features. The original infix representation (\textbf{Figure~\ref{fig:example_postfix_b}}) has issues such as redundancy, semantic sparsity, a high likelihood of illegal transformations, and an overly large search space.

We introduce postfix expressions (\textbf{Figure~\ref{fig:example_postfix_c}}) to solve these problems. 
Postfix expressions don't need many brackets to determine calculation priority. Scanning from left to right suffices to reconstruct the corresponding sequence, greatly reducing sequence-modeling difficulty and computational cost.
They also reduce the ambiguity of the transformation sequence.
Most importantly, it reduces the search space from exponential to a finite set $|C| = |\mathcal{O}| + |\mathcal{F}_0|D + 3$. Here, $|\mathcal{O}|$ represents the operation set size, $|\mathcal{F}_0|$ is the original feature set dimension, $D$ is feature numbers, and $3$ refers to start tokens $<SOS>$, separation token $<SEP>$, and end token $<EOS>$.

\subsection{Problem Statement}
We aim to develop a generative AI system that generates a feature transformation sequence given a tasking dataset. Formally, Let a tabular dataset be $\mathcal{D}=\{(\mathbf{x}^{(i)}, y^{(i)})\}_{i=1}^{N}$ with original feature set $\mathcal{F}_0=\{f_1,\ldots,f_d\}$.
A feature transformation step applies operators $\mathsf{op}$ from the operation set $\mathcal{O}$ on the original features to generate combinations as new features.
The goal is to find the optimal feature transformation sequence $\Gamma^{*}$ that maximizes the downstream ML model $\mathcal{M}$'s performance (i.e.,  balance among accuracy, validity, and stability) on the transformed feature set:
\begin{equation}
\Gamma^{*} = \underset{\Gamma}{\operatorname{argmax}} \mathcal{A}(\mathcal{M}(\text{Transform}(\mathcal{F}_0, \Gamma)), y)
\end{equation}
where $\text{Transform}(\mathcal{F}_0, \Gamma)$ transforms the original feature set $\mathcal{F}_0$ using $\Gamma$, and $\mathcal{A}$ is $\mathcal{M}$'s downstream performance metric.

\begin{figure*}[htbp] 
    \centering 
    \includegraphics[width=0.82\linewidth]{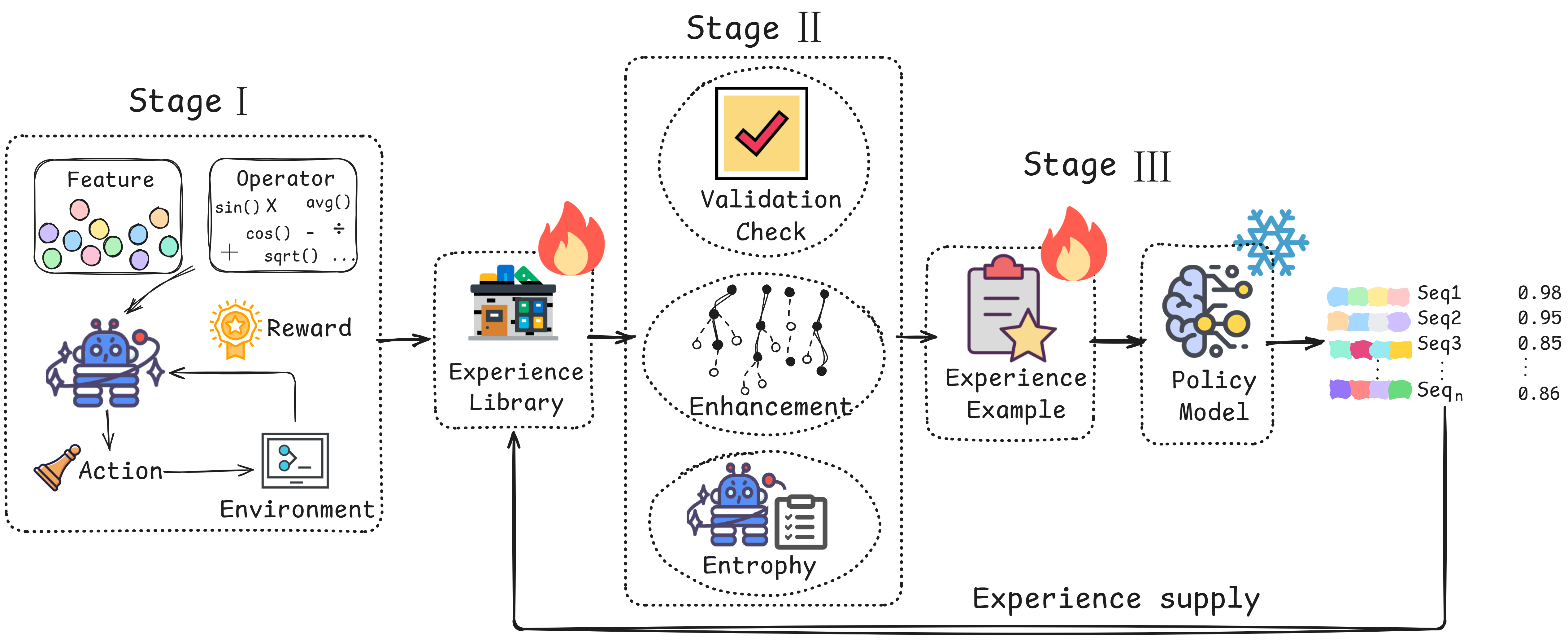} 
    \caption{Data-centric closed-loop optimization of context experiences for LLM-driven feature transformation. Stage I explores high-performing sequences with downstream rewards and stores them in an experience library. Stage II refines the library through validation checks, enhancement, and entropy-based diversity control, and constructs selected experiences into few-shot CoT-style examples. Stage III uses these CoT-style examples to guide the LLM to generate a single transformation sequence, which is then verified and written back to the library.} 
    \label{fig:framework} 
\end{figure*}

\section{Method}
\label{sec:method}

\subsection{Overview of Proposed Method}
To guide general-oriental LLMs generate sequences for data-centric AI tasks. 
We propose a data-centric framework that optimizes \emph{context data} for LLM-based feature transformation.
Instead of treating demonstrations as static prompts, we maintain an evolving experience library and iteratively update it with downstream-verified transformation sequences.
At a high level, our method consists of three stages (\textbf{Figure~\ref{fig:framework}}):
(i) \textbf{exploration}: the reinforcement learning agent explores useful new feature combinations to obtain a verified high-performance demonstration base;
(ii) \textbf{refinement}: we clean, enhance, and diversify these disorganized combination sequences to form sequential trajectories, turning them into reusable experiences that align with the downstream tasks;
and (iii) \textbf{experience usage}: we construct task-specific contexts that guide an LLM to generate improved transformations under downstream evaluation.
This closed loop turns context construction into a principled optimization process aligned with downstream performance.

\subsection{Stage I: RL Exploration for High-performing Sequences}
We first obtain a strong foundation of combination sequences by reinforcement learning.
Our design follows the idea of GRFG~\cite{wang2022group}.
The goal of this stage is clear: we want a set of combination sequences that are \emph{verified} by the downstream task.
These sequences will be stored as initial experiences of the experience library.

A state $s_\ell$ summarizes the current transformed feature set.
An action $a_\ell$ specifies an operator and operand features (i.e., a feature combination).
After we apply $a_\ell$, we obtain a new state $s_{\ell+1}$.
We then evaluate it with the downstream learner.
We define the reward by the improvement of downstream performance:
\begin{equation}
r_\ell = \mathsf{Score}(s_{\ell+1}) - \mathsf{Score}(s_\ell).
\end{equation}

This reward is simple and aligns exploration with the downstream objective.
It also reduces meaningless feature combinations by guiding the model to achieve greater performance improvements.
This is important for later LLM generation.
Without such alignment, an LLM may produce combinations that look good but do not help the task.
In the worst case, it may overfit surface patterns and generate ``pleasing'' but low-quality output (i.e., sycophancy)~\cite{sharma2023towards}.
By using this reward, we keep only task-useful signals.

We store the exploration results as single transformation sequences.
Each sequence is a list of combinations with an associated downstream score.
This format is simple. It is also easy to verify and clean. It fits our goal in this stage: We want to collect reliable building blocks for the experience library.

RL exploration produces a verified set $\mathcal{T}^{\mathrm{RL}}$ that serves as the foundation for the experience library.
Each element in $\mathcal{T}^{\mathrm{RL}}$ is a high-quality sequence, which provides strong, task-aligned examples.

\subsection{Context-as-Data: Experience Library and Context Policy}
Starting from $\mathcal{T}^{\mathrm{RL}}$, we formalize the few-shot context as a data object that can be optimized. We maintain an \emph{experience library} $\mathcal{E}$.
Each experience $e\in\mathcal{E}$ is a downstream-verified demonstration, which contains:
(i) a transformation \emph{sequence} $\Gamma$ consisting of step-wise feature combinations,
and (ii) final downstream scores $m_L=\mathsf{Score}(\mathcal{F}(\Gamma))$.
Here $\Gamma$ is a sequence of valid combinations.
In Stage II, we will further refine and reorganize these sequences into performance-aware trajectories.

We denote an experience as
\begin{equation}
e=\big(\Gamma=[\tau_1,\ldots,\tau_L],\; m_L\big),
\end{equation}
where $\tau_\ell$ is the $\ell$-th feature combination (a transformation step), and $\Gamma$ denotes a sequence of combinations.
This form is simple, making each experience executable and tying each sequence to a downstream performance score.
So the library can store reliable examples for later reuse.

At each iteration, we build a small few-shot context from the experience library $\mathcal{E}$ and use it to guide the LLM.
The selection prefers experiences with strong downstream performance and avoids highly redundant ones.
This makes the selection process \emph{data-centric}: it is driven by downstream scores and updated by the refinement procedure in Stage II, together with the diversity control in Sec.~\ref{sec:entropy}.
As a result, the context evolves over iterations as reusable token-level knowledge, which improves generation quality and better aligns the LLM with the downstream task.

\subsection{Stage II: Three-level Refinement for Few-shot Context Construction}
Stage I provides initial sequences and their final scores, and we store them as raw experiences in $\mathcal{E}$.
In Stage II, we clean and enrich $\mathcal{E}$ to build the context input for the LLM.
Our refinement follows a simple process: we check each combination and sequence, reorganize sequences into trajectories (CoT-style), and reduce redundancy with entropy-guided selection.

\subsubsection{Sequence Validation Check (Local Reliability)}
We first ensure each feature combination is correct.
For each candidate combination $\tau$, we apply a checker to remove invalid or harmful transformations:
\begin{itemize}[leftmargin=*]
    \item \textbf{Syntactic/typing validity:} every referenced feature exists, and each operator is applicable to the feature types.
    \item \textbf{Numerical stability:} avoid division by near-zero, invalid $\log/\sqrt{\cdot}$ inputs, overflow, and excessive NaNs.
    \item \textbf{Minimum utility:} discard combinations that consistently lead to negative gains under cross-validation.
\end{itemize}

Only combinations passing $\mathsf{Check}$ are kept. A sequence is valid only if all its combinations are valid and the whole sequence is executable.
This step reduces the invalid rate and prevents noisy experiences from entering the few-shot context.

\subsubsection{CoT Trajectory Construction and Enhancement}
Next, we reorganize verified sequences into performance-aware trajectories to form CoT-style demonstrations.
The goal is to show the LLM a clear evolution path: \textbf{how transformations can be improved to push downstream performance upward.}
Specifically, we take multiple verified sequences and order them by downstream performance.
This produces a trajectory whose steps reflect ``improvement over time''.
Compared with isolated examples, such trajectories provide stronger guidance aligned with downstream tasks.
This method shows LLMs a Chain of Improvement to encourage them generating sequence with higher downstream performance.

\begin{figure}[htbp] 
    \centering 
    \includegraphics[width=0.95\linewidth]{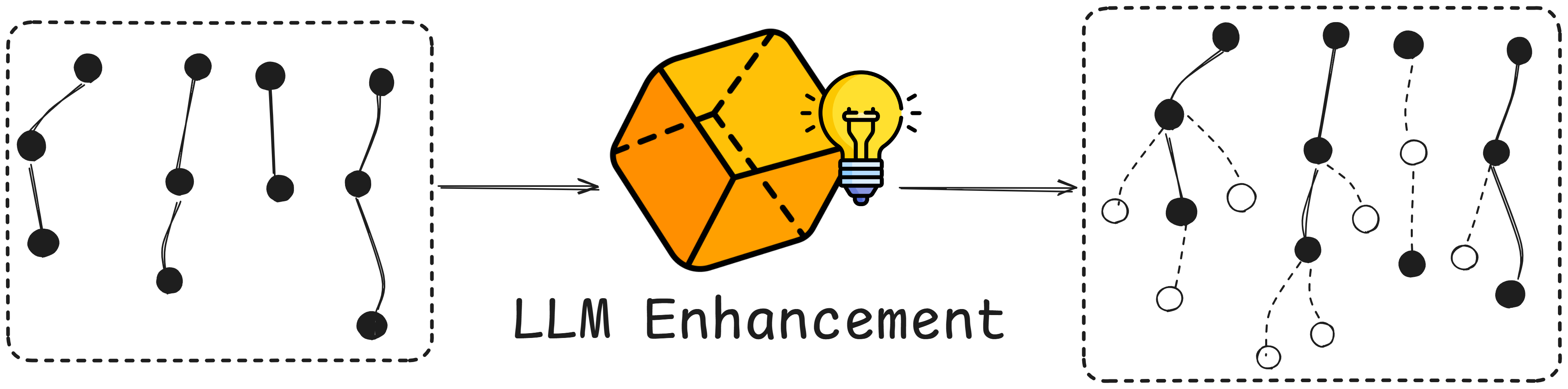} 
    \caption{LLM-based enhancement for CoT trajectory construction. Starting from verified high-quality sequences (left), an LLM proposes gap-filling and local variants to enrich the neighborhood of transformation patterns, producing more complete and diverse CoT-style trajectories (right).} 
    \label{fig:enhancement} 
\end{figure}

We further apply enhancement using an LLM (\textbf{Figure~\ref{fig:enhancement}}).
In practice, some useful steps are missing between two strong sequences.
The LLM proposes intermediate or nearby variants to fill these gaps.
All enhanced candidates must pass the same combination/sequence checks and downstream verification before being added back to $\mathcal{E}$.
This makes the CoT path more complete and improves the library quality for later context selection.

\subsubsection{Entropy-guided Diversity Selection (Coverage vs. Redundancy)}
\label{sec:entropy}
Experiences may collapse into a narrow pattern family.
To ensure broad coverage of the transformation space with low redundancy, we compute a discrete \emph{signature} $z(e)$ for each experience $e$ to summarize its structural pattern.
Let $p(z)$ be the empirical distribution of signatures in a selected set $\mathcal{S}$; we measure coverage by entropy:
\begin{equation}
H(\mathcal{S}) = -\sum_{z} p(z)\log p(z).
\end{equation}
We measure redundancy by average pairwise similarity $\mathrm{sim}(e_i,e_j)$ within $\mathcal{S}$:
\begin{equation}
\mathrm{Red}(\mathcal{S}) = \frac{1}{|\mathcal{S}|(|\mathcal{S}|-1)} \sum_{i\neq j}\mathrm{sim}(e_i,e_j).
\end{equation}
We select contexts by optimizing a quality--diversity trade-off:
\begin{equation}
\max_{\mathcal{S}\subset\mathcal{E},\,|\mathcal{S}|=K}
\;\;\underbrace{\mathbb{E}_{e\in\mathcal{S}}[\mathsf{Score}(e)]}_{\text{quality}}
+ \lambda \underbrace{H(\mathcal{S})}_{\text{coverage}}
- \mu \underbrace{\mathrm{Red}(\mathcal{S})}_{\text{redundancy}},
\end{equation}
which can be solved effectively through greedy selection.
This entropy-guided step is used when selecting sequences for the few-shot context, so that we keep coverage high and redundancy low.

\subsection{Stage III: Experience-conditioned Sequence Generation and Write-back}
Based on an evolving experience library $\mathcal{E}$ and a few-shot context, Stage III uses the context to guide the LLM to generate new sequences, and then updates $\mathcal{E}$ with newly verified sequences.

\subsubsection{Context construction}
The few-shot context is mainly CoT-style demonstrations constructed in Stage II.
On top of these demonstrations, we add lightweight guidance to help the LLM generate valid and task-useful sequences (e.g., allowed operators $\Omega$, feature references, and a simple output schema).
We use CoT-style demonstrations as guidance, but we only keep the final sequence output to simplify execution and verification.

\subsubsection{LLM generation as a single sequence}
The LLM generates a set of candidate sequences
$\hat{\mathcal{G}}_t=\{\hat{\Gamma}_j\}_{j=1}^{B}$.
Each $\hat{\Gamma}_j$ is a sequence of feature combinations, i.e., $\hat{\Gamma}_j=[\hat{\tau}_1,\ldots,\hat{\tau}_{L_j}]$.
We restrict generation with simple rules.
For example, we only allow operators in $\Omega$, and we limit the maximum length.

\subsubsection{Downstream verification and selection}
We verify each generated sequence before writing back.
We first apply $\mathsf{Check}(\cdot)$ to filter invalid combinations and invalid sequences.
Then we execute the sequence to obtain transformed features and evaluate it with the downstream learner:
\begin{equation}
\hat{m}_L(\hat{\Gamma})=\mathsf{Score}\big(\mathcal{M};\mathcal{D}\text{ with }\mathcal{F}(\hat{\Gamma})\big).
\end{equation}
We rank candidates by $\hat{m}_L(\hat{\Gamma})$ and keep the best ones.
We also remove near-duplicate candidates to reduce redundancy.

\subsubsection{Experience write-back (closed-loop update)}
For each selected sequence $\hat{\Gamma}$, we create a new experience
$\hat{e}=(\phi(\mathcal{D}),\hat{\Gamma},\hat{m}_L(\hat{\Gamma}))$
and merge it into the library:
\begin{equation}
\mathcal{E}_{t+1} \leftarrow \mathcal{E}_t \cup \hat{\mathcal{G}}^{\star}_t,
\end{equation}
where $\hat{\mathcal{G}}^{\star}_t$ denotes the verified top candidates at iteration $t$.
This update is driven by downstream scores.
So the library gradually shifts toward sequences that improve the task.
Over iterations, the selected few-shot context becomes stronger and more diverse.

\begin{table*}[htbp]
  \begin{center}
    \caption{Overall downstream performance comparison. We report F1 for classification and $1$-RAE for regression (higher is better), averaged over 5-fold cross-validation. Best results are in bold and second-best results are underlined.}
    \label{tab:overallresult}
    \resizebox{\textwidth}{!}{\begin{tabular}{lcccccccccccccccccccc}
      \toprule
      \textbf{Dataset} & \textbf{Source} & \textbf{Task} & \textbf{Samples} & \textbf{Features} &
      \textbf{Original} & \textbf{RDG} & \textbf{PCA} & \textbf{LDA} & \textbf{ERG} & \textbf{AFAT} & \textbf{AutoFeat} & \textbf{NFS} & \textbf{TTG} & \textbf{GRFG} & \textbf{MOAT} & \textbf{OpenFE} & \textbf{CAAFE} & \textbf{FeatLLM} & \textbf{ELLM-FT} & \textbf{Ours} \\
      \midrule
      Amazon Employee & Kaggle & C & 32,769 & 9 &
      93.37\% & 92.31\% & 92.29\% & 91.64\% & 92.43\% & 92.97\% & 93.29\% & 93.21\% & 92.79\% & 93.02\% & 93.13\% & 93.44\% & 91.41\% & \ul{93.62\%} & 93.17\% & \textbf{94.41\%} \\
      German Credit & UCIrvine & C & 1,000 & 24 &
      74.20\% & 68.01\% & 67.92\% & 63.91\% & 74.43\% & 68.32\% & 74.86\% & 68.67\% & 64.51\% & 68.29\% & 72.44\% & 74.50\% & 59.92\% & 76.35\% & \ul{76.39\%} & \textbf{85.32\%} \\
      Higgs Boson & UCIrvine & C & 50,000 & 28 &
      69.66\% & 67.51\% & 53.45\% & 51.32\% & 69.02\% & 69.70\% & 67.35\% & 69.17\% & 68.99\% & 69.77\% & 69.12\% & 69.66\% & 61.26\% & \ul{70.35\%} & 69.66\% & \textbf{72.29\%} \\
      Ionosphere & UCIrvine & C & 351 & 34 &
      93.37\% & 91.17\% & 92.87\% & 65.53\% & 92.02\% & 92.87\% & 93.37\% & 91.17\% & 90.31\% & 93.16\% & 95.69\% & 93.37\% & 92.84\% & 95.38\% & \ul{96.01\%} & \textbf{97.14\%} \\
      Lymphography & UCIrvine & C & 148 & 18 &
      83.19\% & 79.36\% & 70.38\% & 70.38\% & 83.73\% & 82.38\% & 79.26\% & 85.25\% & 82.38\% & 85.51\% & 88.38\% & 83.73\% & 75.00\% & 85.24\% & \ul{90.54\%} & \textbf{95.07\%} \\
      Messidor Feature & UCIrvine & C & 1,151 & 19 &
      69.09\% & 62.38\% & 67.21\% & 47.52\% & 66.90\% & 66.55\% & 69.08\% & 63.77\% & 66.46\% & 69.24\% & 73.02\% & 69.09\% & 66.10\% & 72.62\% & \ul{74.80\%} & \textbf{76.98\%} \\
      PimaIndian & Kaggle & C & 768 & 8 &
      80.68\% & 76.04\% & 63.80\% & 63.80\% & 76.17\% & 76.56\% & 80.86\% & 74.87\% & 74.48\% & 75.39\% & 80.73\% & 80.86\% & 79.86\% & 89.66\% & \ul{89.66\%} & \textbf{93.29\%} \\
      Spam Base & UCIrvine & C & 4,601 & 57 &
      94.53\% & 90.61\% & 81.66\% & 88.89\% & 91.70\% & 91.20\% & 94.54\% & 92.50\% & 91.91\% & 92.20\% & 92.90\% & 94.53\% & 88.51\% & 95.03\% & \textbf{96.68\%} & \ul{96.19\%} \\
      SpectF & UCIrvine & C & 267 & 44 &
      76.06\% & 76.03\% & 70.92\% & 66.29\% & 75.66\% & 76.03\% & 76.06\% & 79.40\% & 76.03\% & 81.65\% & \ul{86.95\%} & 76.06\% & 70.60\% & 80.07\% & 86.14\% & \textbf{87.16\%} \\
      SVMGuide3 & LibSVM & C & 1,243 & 21 &
      81.85\% & 78.68\% & 67.60\% & 65.24\% & 82.62\% & 79.49\% & \ul{83.05\%} & 79.16\% & 79.81\% & 81.17\% & 81.74\% & 81.85\% & 75.30\% & 82.54\% & 82.70\% & \textbf{87.68\%} \\
      UCI Credit & UCIrvine & C & 30,000 & 23 &
      79.29\% & 80.32\% & 73.27\% & 74.37\% & 80.16\% & 80.32\% & 79.72\% & 80.13\% & 79.81\% & 80.67\% & \ul{80.87\%} & 80.11\% & 76.80\% & 76.39\% & 79.29\% & \textbf{80.88\%} \\
      Wine Quality Red & UCIrvine & C & 999 & 11 &
      60.95\% & 46.65\% & 42.21\% & 43.31\% & 46.10\% & 48.05\% & 62.52\% & 46.21\% & 46.71\% & 47.01\% & 62.10\% & 53.71\% & 51.74\% & \ul{62.65\%} & 61.11\% & \textbf{68.59\%} \\
      Wine Quality White & UCIrvine & C & 4,898 & 11 &
      54.75\% & 52.41\% & 43.01\% & 44.94\% & 51.04\% & 51.67\% & 54.26\% & 52.51\% & 53.12\% & 53.41\% & 54.52\% & 54.75\% & 42.82\% & \ul{56.87\%} & 55.03\% & \textbf{66.95\%} \\
      \midrule
      Airfoil & UCIrvine & R & 1,503 & 5 &
      0.5749 & 0.5193 & 0.2730 & 0.2201 & 0.5193 & 0.5210 & 0.5746 & 0.5193 & 0.5003 & 0.5587 & 0.5967 & 0.5746 & N/A & 0.5877 & \ul{0.6174} & \textbf{0.7594} \\
      Housing Boston & Kaggle & R & 506 & 13 &
      0.4148 & 0.4043 & 0.1048 & 0.0201 & 0.4090 & 0.4161 & 0.4149 & 0.4251 & 0.3967 & 0.4043 & 0.4463 & 0.4148 & N/A & 0.4442 & \ul{0.4564} & \textbf{0.7295} \\
      Openml 586 & OpenML & R & 1,000 & 25 &
      0.6311 & 0.5681 & 0.1109 & 0.1109 & 0.6147 & 0.5435 & 0.6329 & 0.5443 & 0.5443 & 0.5768 & 0.6251 & 0.6311 & N/A & \ul{0.6477} & 0.6328 & \textbf{0.7406} \\
      Openml 589 & OpenML & R & 1,000 & 25 &
      0.5388 & 0.5091 & 0.0112 & 0.0112 & 0.5103 & 0.5087 & 0.5423 & 0.5053 & 0.5032 & 0.5047 & 0.5139 & 0.5388 & N/A & 0.5545 & \ul{0.5836} & \textbf{0.6602} \\
      Openml 607 & OpenML & R & 1,000 & 50 &
      0.6207 & 0.5208 & 0.1071 & 0.1071 & 0.5553 & 0.5158 & 0.6191 & 0.5194 & 0.5222 & 0.6021 & 0.6051 & \ul{0.6207} & N/A & 0.5608 & 0.6089 & \textbf{0.7408} \\
      Openml 616 & OpenML & R & 500 & 50 &
      0.3736 & 0.0701 & 0.0242 & 0.0241 & 0.1937 & 0.1489 & 0.3924 & 0.1667 & 0.1567 & 0.3722 & 0.4063 & 0.3736 & N/A & 0.3836 & \ul{0.4082} & \textbf{0.5789} \\
      Openml 618 & OpenML & R & 1,000 & 50 &
      0.4402 & 0.3720 & 0.1016 & 0.0521 & 0.3561 & 0.2472 & 0.4407 & 0.3473 & 0.3467 & 0.4562 & \ul{0.4734} & 0.4402 & N/A & 0.4597 & 0.4734 & \textbf{0.6546} \\
      Openml 620 & OpenML & R & 1,000 & 25 &
      0.6434 & 0.5111 & 0.1138 & 0.0293 & 0.5466 & 0.5267 & \ul{0.6576} & 0.5130 & 0.5123 & 0.5591 & 0.5722 & 0.6434 & N/A & 0.5725 & 0.6203 & \textbf{0.6925} \\
      Openml 637 & OpenML & R & 500 & 50 &
      0.3162 & 0.1364 & 0.0352 & 0.0433 & 0.1521 & 0.1758 & \ul{0.3251} & 0.1521 & 0.1439 & 0.2071 & 0.2125 & 0.3162 & N/A & 0.2945 & 0.2946 & \textbf{0.5471} \\
      \bottomrule
    \end{tabular}}
  \end{center}
\end{table*}

\section{Experiments}
\label{sec:experiments}

We design experiments to answer the following questions:
\textbf{(Q1)} Does our data-centric experience evolution framework improve downstream performance over baselines on tabular datasets?
\textbf{(Q2)} Does the closed-loop write-back bring consistent gains compared with one-shot LLM generation?
\textbf{(Q3)} How much does each stage contribute (RL exploration, three-level refinement, and context usage), and is CoT-style organization necessary?
\textbf{(Q4)} Is the method transferable across different LLMs (API-based and open-source) and robust to downstream models and data shifts?
\textbf{(Q5)} What are the cost--performance trade-offs, and what patterns can we observe from LLM behaviors during FT?

\subsection{Experimental Setup}
\label{subsec:setup}

\noindent\textbf{Datasets.}
We evaluate on a diverse collection of tabular benchmarks from UCIrvine~\cite{uci_dataset_2023}, CPLM~\cite{CPLM_2023}, Kaggle~\cite{howard_kaggle_2023}, and OpenML~\cite{Openml_dataset_2023}.
These datasets cover both classification and regression tasks. The corresponding statistics (feature and sample numbers) and task types are presented in \textbf{Table~\ref{tab:overallresult}}, where 'C' represents classification and 'R' represents regression.
For all datasets and all methods, we use the same 5-fold cross-validation protocol and report results averaged over the five folds.

\noindent\textbf{Downstream evaluator.}
We evaluate feature transformation by training a fixed downstream model on the transformed feature set and measuring its predictive performance.
We use the same downstream model and the same hyperparameter setting for all methods to ensure fairness.
For classification tasks, we report the F1-score:
$F1 = 2 \cdot \frac{\text{Precision} \cdot \text{Recall}}{\text{Precision} + \text{Recall}}$, where $\text{Precision}=\frac{TP}{TP+FP}$ and $\text{Recall}=\frac{TP}{TP+FN}$.
For multi-class datasets, we use macro-F1 by averaging over classes.
For regression tasks, we report $1$-RAE (Relative Absolute Error), where a larger value indicates better performance: $1\text{-RAE} = 1 - \frac{\|\bm{y}_{pred} - \bm{y}_{real}\|_1}{\|\bm{y}_{real} - \bar{\bm{y}}_{real}\|_1}$, where $\bm{y}_{pred}$ is the predicted value, $\bm{y}_{real}$ is the true value, and $\bar{\bm{y}}_{real}$ is the mean of $\bm{y}_{real}$.

\noindent\textbf{Operator Set.}
We use a fixed operator set $\mathcal{O}$ that contains unary operators $\mathcal{O}_1$ and binary operators $\mathcal{O}_2$.
The unary set is
$\mathcal{O}_1=\{\texttt{sqrt}, \\ \texttt{square},\texttt{cube},\texttt{reciprocal},\texttt{log},\texttt{sin},\texttt{cos},\texttt{tanh},\texttt{sigmoid},\texttt{standard}, \\ \texttt{normalize},\texttt{quantile}\}$,
and the binary set is $\mathcal{O}_2=\{+,-,\times,\div\}$.
Unary operators are suited for a single feature, and binary operators combine two features.
All baselines that support operator control are restricted to the same $\mathcal{O}$.

\noindent\textbf{LLM Settings.}
We evaluate nine policy LLMs in Stage III, including Llama-3.2-3B, Llama-3.1-8B, Llama-4, GPT-4o, o1, o3-mini, DeepSeek-V3, Claude-3.7, and Qwen3-4B.
To ensure a fair comparison, we use a unified generation protocol and keep the sampling hyperparameters as consistent as possible across models.
Specifically, we set $\texttt{temperature}=0.7$, $\texttt{top\_p}=0.9$, and $\texttt{max\_new\_tokens} = 500$ for all LLMs.
For open-source models that support top-$k$ sampling, we set $\texttt{top\_k}=50$.
We fix the budget to $T=10$ iterations and evaluate $B=10$ candidates per iteration, resulting in $TB=100$ calls per dataset.

\noindent\textbf{Baselines.}
We compare our method with representative baselines from three families (see \textbf{Table~\ref{tab:overallresult}}).
\textbf{(i)} Random/linear baselines: RDG, PCA~\cite{mackiewicz1993principal}, and LDA.
\textbf{(ii)} Classical automated feature engineering and search-based FT: ERG, AFAT~\cite{horn2020autofeat}, AutoFeat~\cite{horn2019autofeat}, NFS~\cite{chen2019neural}, TTG~\cite{khurana2018feature}, GRFG~\cite{wang2022group}, MOAT~\cite{wang2023reinforcement}, and OpenFE~\cite{zhang2023openfe}.
\textbf{}(iii) LLM-based feature engineering: FeatLLM~\cite{han2024large}, CAAFE~\cite{hollmann2023large}, and ELLM-FT~\cite{gong2025evolutionary}.

\noindent\textbf{Configurations.}
All experiments were conducted on the Ubuntu 22.04.3 LTS operating system, with a 13th-generation Intel(R) Core(TM) i9-13900KF CPU and an NVIDIA GeForce RTX 4090 GPU. The experiments were conducted using Python 3.11.5 and PyTorch 2.0.1.

\subsection{Benchmark Performance}
To check whether our method works well on the feature transformation task, we compare our method with feature transformation baselines on various tabular benchmarks.
All methods share the same operator set $\mathcal{O}$ and use the same downstream model and 5-fold cross-validation protocol.
For each dataset, we report the average performance over folds.

\textbf{Table~\ref{tab:overallresult}} shows the results on both classification and regression tasks.
Our method achieves the best average ranking.
It outperforms classical search-based FT methods (e.g., GRFG/MOAT) and automated feature engineering pipelines (e.g., AFAT/AutoFeat).
Compared with other LLM-based baselines (CAAFE/FeatLLM/ELLM-FT), our method is also more stable.
In many cases, the gains are more obvious in datasets where a naive search or fixed prompting is likely to produce invalid or redundant transformations. Our library-based context provides verified and reusable signals.
We also observe that strong classical methods remain competitive, which suggests that pure operator application can still help when the task is simple or the baseline feature space is already strong.

\textbf{Insight.}
These results support our main claim: for LLM-driven feature transformation, optimizing the \emph{context data} is a practical and effective data-centric strategy.
Stage I provides task-aligned exploration signals through downstream rewards.
Stage II converts raw sequences into higher-quality CoT-style demonstrations with verification, enhancement, and redundancy control.
Stage III uses these demonstrations to guide the LLM to output a single high-utility sequence.
Together, the closed-loop turns feature transformation into an experience-evolution process, improving downstream alignment and reducing meaningless combinations.

\subsection{Closed-loop Gains over One-shot Generation}
We compare our closed-loop experience evolution with one-shot LLM generation under the same evaluation budget.
The one-shot baseline generates candidates without updating the experience library.
We consider two one-shot settings: \emph{fixed} uses the same few-shot context for all calls, and \emph{re-sample} re-selects the few-shot context from the library at each call.
In contrast, our closed-loop method writes back verified high-performing sequences to update the library over iterations.

\begin{figure}[htbp]
    \centering
    \begin{subfigure}[t]{0.23\textwidth}
        \centering
        \includegraphics[width=\linewidth]{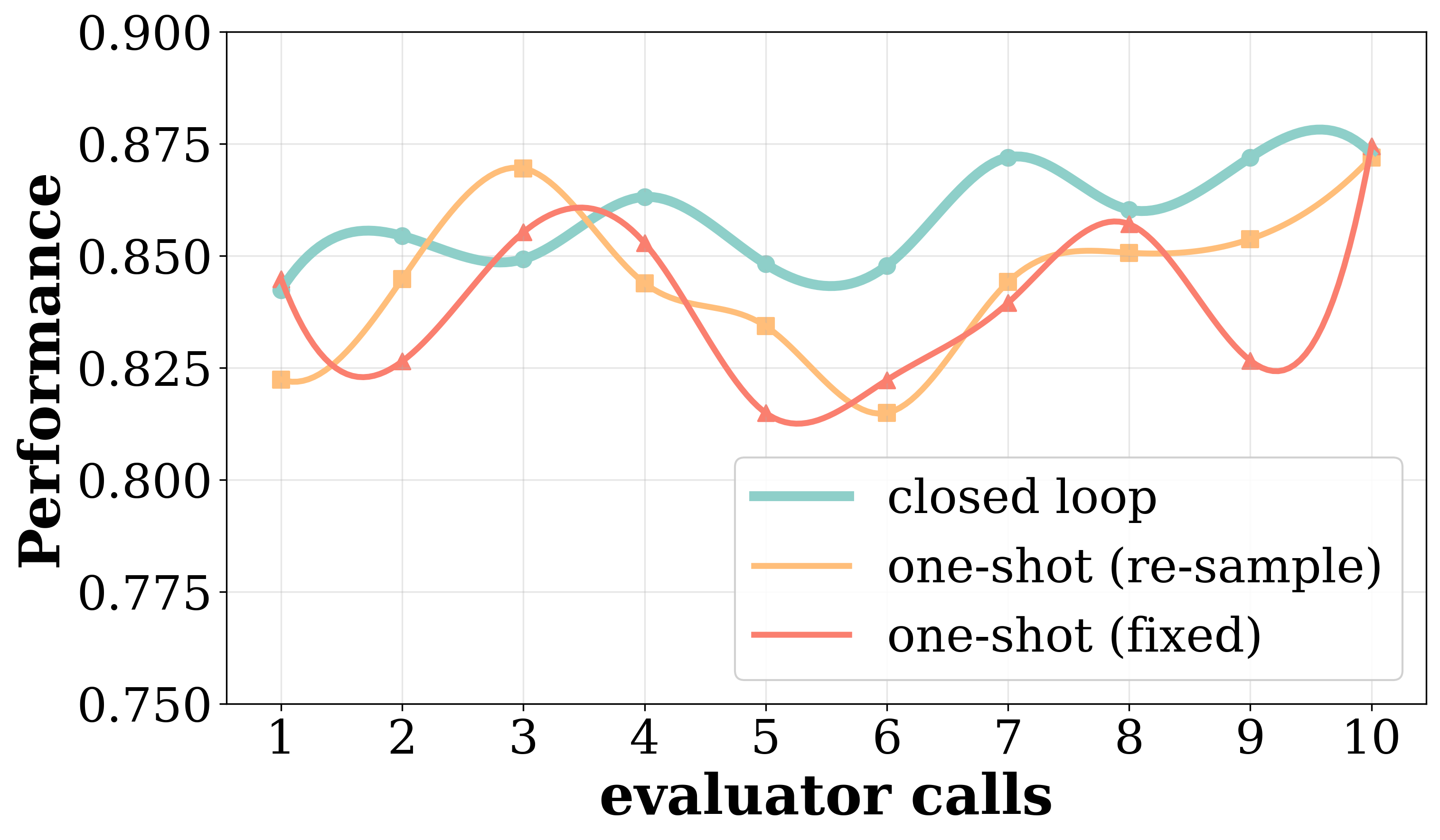}
        \caption{\textbf{SpectF } (C).}
        \label{fig:closedloop_ds1}
    \end{subfigure}
    \hfill
    \begin{subfigure}[t]{0.23\textwidth}
        \centering
        \includegraphics[width=\linewidth]{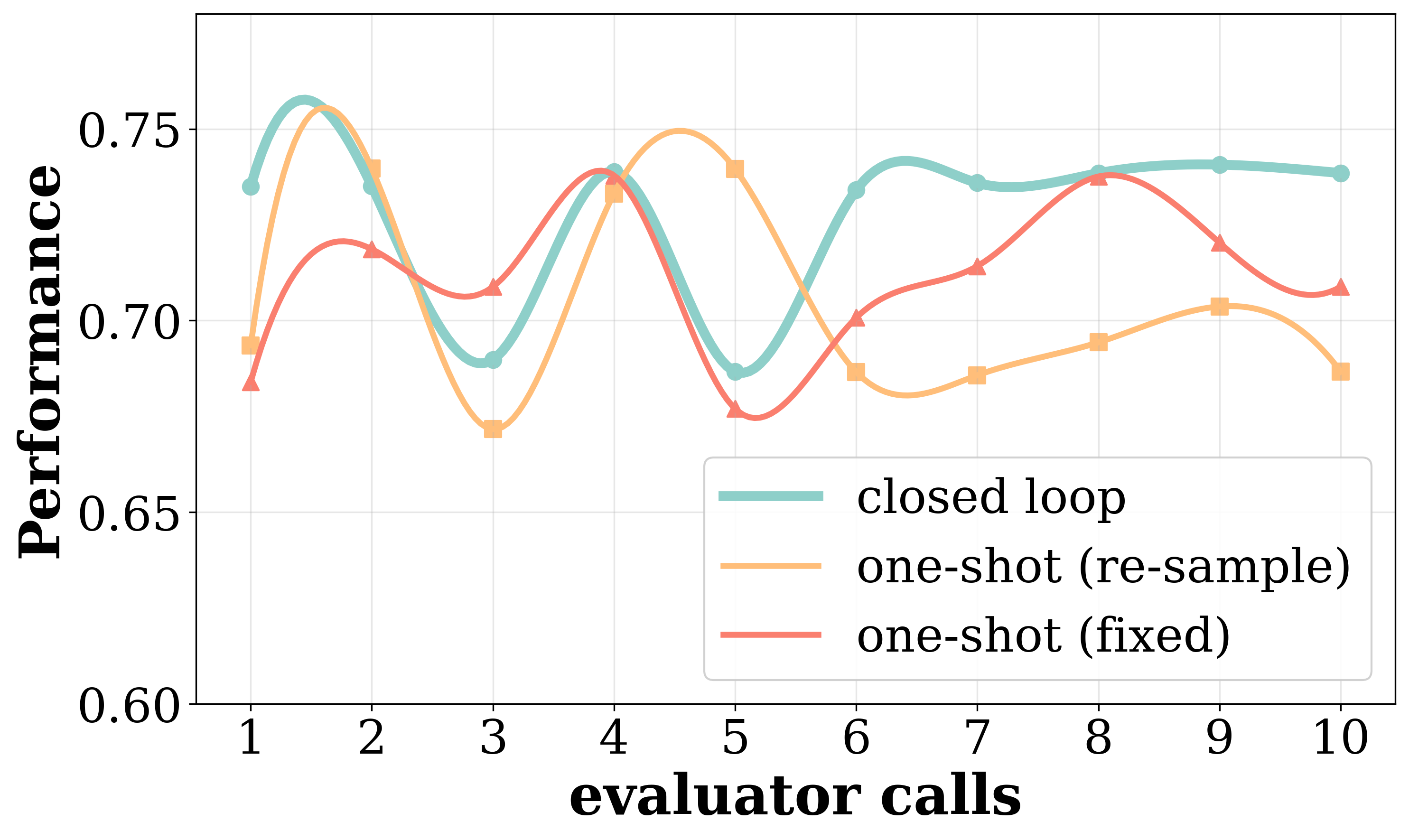}
        \caption{\textbf{OpenML 586} (R).}
        \label{fig:closedloop_ds2}
    \end{subfigure}
    \caption{\textbf{Closed-loop vs. one-shot generation} on two datasets under the same evaluation budget. Each plot reports downstream performance on different evaluator calls.}
    \label{fig:closed_loop}
\end{figure}

\textbf{Figure~\ref{fig:closed_loop}} reports the downstream performance on the evaluator calls in two datasets.
One-shot generation is unstable: even with context re-sampling, the performance fluctuates and does not show consistent improvement.
In contrast, the closed-loop method is more stable and shows a clearer improvement trend as evaluator calls increase, leading to better final performance.
This suggests that write-back updates provide benefits beyond stochastic sampling and static prompting.

\textbf{Insight.}
This study supports our claim of context-as-data.
CoT-style demonstrations provide strong initial guidance, but the key advantage comes from closed-loop updates: verified sequences are distilled into reusable experiences, which improves the context quality and prevents the LLM from redundant or low-utility combinations.
As the library evolves, generation becomes more stable and better aligned with the downstream task.

\subsection{Ablation of Three-level Refinement}
To show the effect of each component in stage II, we evaluated variants that remove combination-level validity checking (w/o combination check), remove CoT-style organization and keep list-only examples (w/o CoT), remove enhancement (w/o enhancement), and replace entropy-guided selection with naive context selection (w/o entropy, top-$K$ or random-$K$).

\begin{table}[htbp]
\centering
\caption{Ablation of Stage II three-level refinement on two representative datasets. We report downstream performance (higher is better) and error ratio (lower is better).}
\label{tab:ablation_stage2}
\resizebox{\linewidth}{!}{
\begin{tabular}{l|cc|cc}
\toprule
\multirow{2}{*}{\textbf{Variant}} 
& \multicolumn{2}{c|}{\textbf{SpectF}} 
& \multicolumn{2}{c}{\textbf{OpenML 586}} \\
\cmidrule(lr){2-3}\cmidrule(lr){4-5}
& \textbf{Perf.} $\uparrow$ & \textbf{Error} $\downarrow$
& \textbf{Perf.} $\uparrow$ & \textbf{Error} $\downarrow$ \\
\midrule
\textbf{Full (Ours)} 
& 87.16\% & 2.70\% & 0.7406 & 10.97\% \\
w/o combination check 
& 87.16\% & 4.23\% & 0.7406 & 23.70\% \\
w/o CoT (list-only) 
& 80.67\% & 1.45\% & 0.6898 & 5.55\% \\
w/o enhancement 
& 85.43\% & 1.75\% & 0.6943 & 5.36\% \\
w/o entropy (top-$K$) 
& 86.67\% & 2.74\% & 0.7349 & 11.76\% \\
w/o entropy (random-$K$) 
& 86.59\% & 2.70\% & 0.7203 & 11.11\% \\
\bottomrule
\end{tabular}}
\end{table}

\begin{figure}[!b]
    \centering
    \begin{subfigure}[t]{\linewidth}
        \centering
        \includegraphics[width=0.8\linewidth]{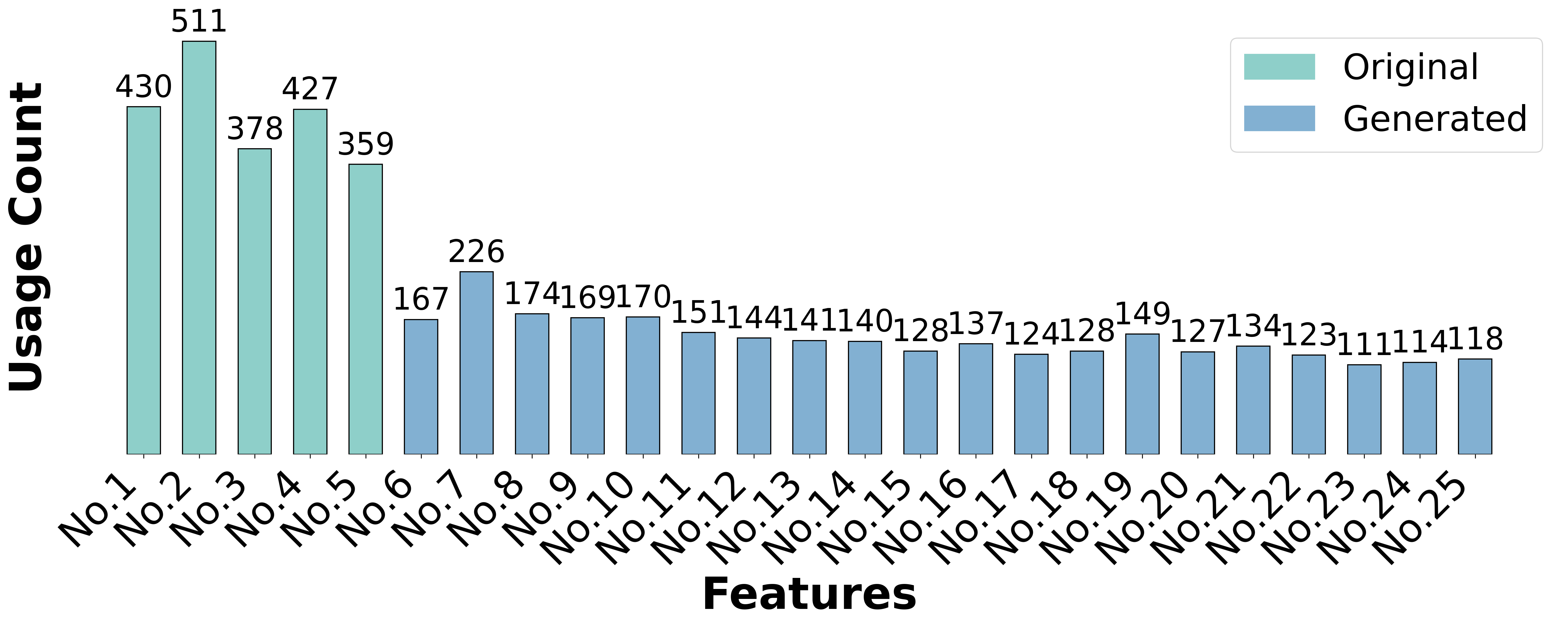}
        \caption{\textbf{Before refinement.}}
        \label{fig:diversity_before}
    \end{subfigure}

    \vspace{0.6em}

    \begin{subfigure}[t]{\linewidth}
        \centering
        \includegraphics[width=0.8\linewidth]{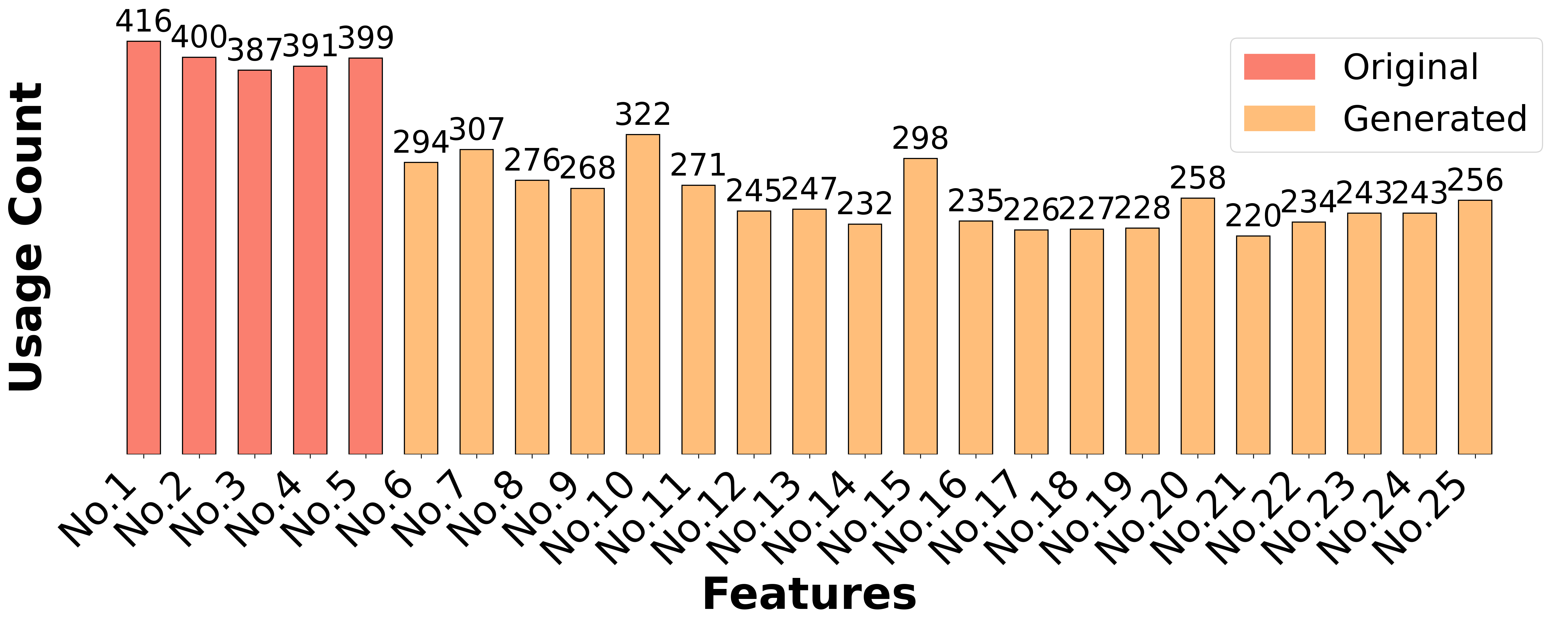}
        \caption{\textbf{After refinement.}}
        \label{fig:diversity_after}
    \end{subfigure}

    \caption{\textbf{Feature usage before vs. after the refinement.}}
    \label{fig:feature_usage_refine}
\end{figure}

\textbf{Table~\ref{tab:ablation_stage2}} reports the final downstream performance and error ratio.
Removing combination-check keeps performance similar in this setting but increases the error ratio, especially on OpenML 586 (10.97\% $\rightarrow$ 23.70\%), showing that local filtering is important for preventing invalid or non-executable combinations from entering the experience context.
Removing CoT structure leads to the largest performance drop on both datasets (e.g., 87.16\% $\rightarrow$ 80.67\% on SpectF), indicating that CoT-style demonstrations provide essential guidance for composing feature combinations toward higher downstream scores.
The enhancement brings gains over the non-enhanced variant, suggesting that LLM-proposed neighborhood variants complement RL explorations by filling missing but useful patterns.
Finally, replacing entropy-guided selection with naive top-$K$ or random-$K$ slightly degrades performance and increases errors, showing that context selection quality matters.

\textbf{Figure~\ref{fig:feature_usage_refine}} visualizes how refinement changes the experience distribution.
Before refinement, feature usage is unbalanced: a small subset of features dominates the generated sequences.
After refinement, the usage distribution becomes more balanced across features, indicating improved coverage and reduced redundancy.
This aligns with the goal of entropy-guided selection and CoT-style construction: the refined context is not only cleaner but also more diverse, and then more informative for LLM generation.

\textbf{Insight.} These results support our data-centric view.
The three-level refinement reduces redundancy and improves coverage: checking improves reliability by reducing invalid generations, CoT-style organization provides stronger compositional guidance, and enhancement plus diversity-aware selection prevent the context from collapsing into redundant patterns.
Together, the three-level refinement turns raw sequences into a higher-quality and better-covered experience context, which leads to more stable and better-aligned feature transformation.

\subsection{Ablation of Initial Experience Library Size}

We study how the size of the \emph{initial} RL experience library affects the final performance.
Specifically, we change the number of explored sequences used to initialize the experience library, i.e., $|\mathcal{E}_0|\in\{1,5,20,50,100\}$.

\begin{figure}[htbp]
    \centering
    \includegraphics[width=0.9\linewidth]{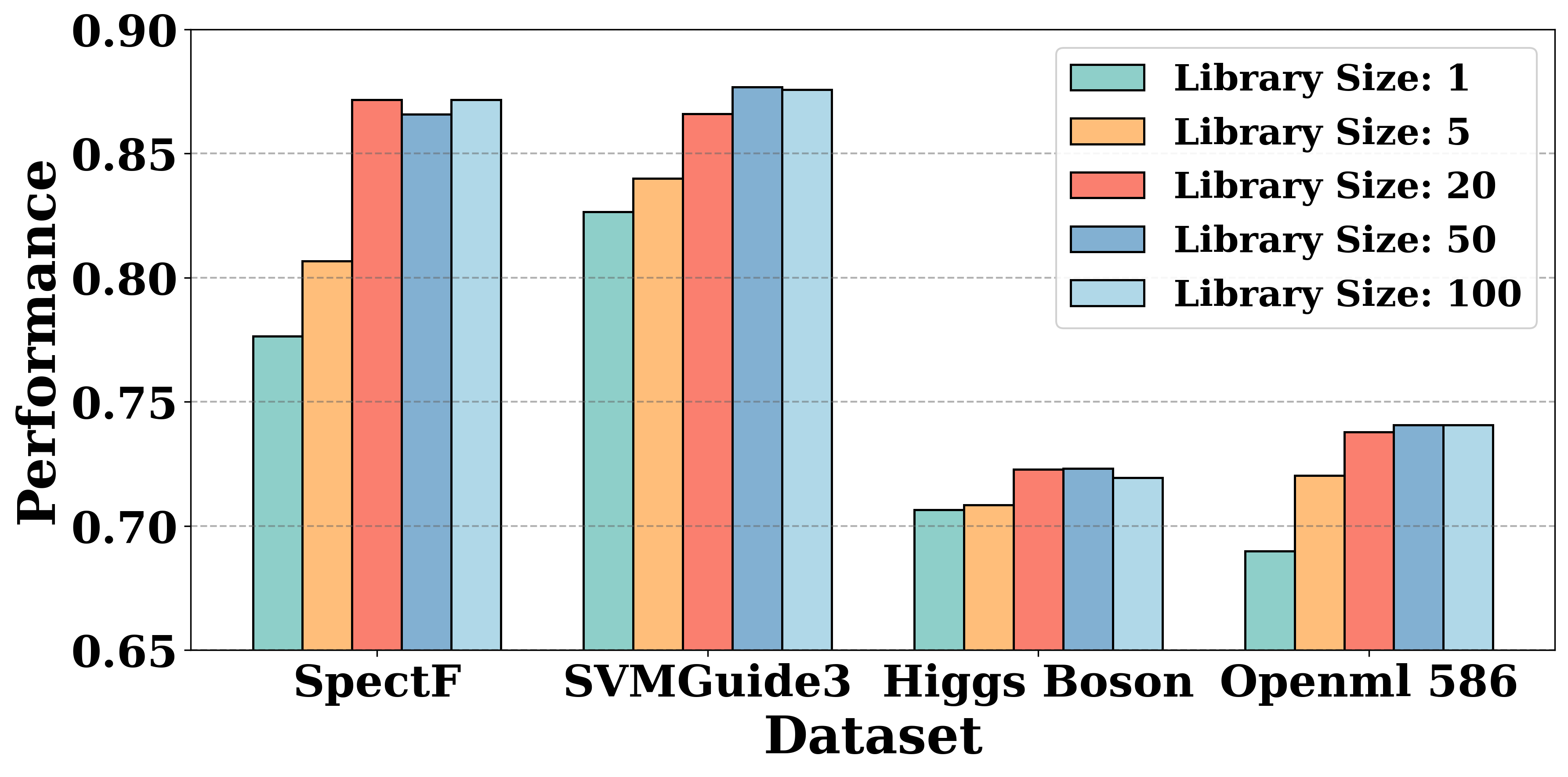}
    \caption{\textbf{Effect of initial experience library size.}
    We change the initial library size $|\mathcal{E}_0|\in\{1,5,20,50,100\}$ (number of RL explored sequences) and report the final downstream performance on four representative datasets.}

    \label{fig:library_size}
\end{figure}

\textbf{Figure~\ref{fig:library_size}} shows that performance improves substantially when $|\mathcal{E}_0|$ increases from 1 to 20.
Once $|\mathcal{E}_0|\ge 20$, the performance becomes more stable, and the gains show diminishing returns on all four datasets.
This suggests that a moderate number of RL explorations is sufficient to provide a strong starting point, while further enlarging the library brings only marginal additional benefit.

\textbf{Insight.}
This ablation supports the role of Stage I as a \emph{seed generator} for context data.
A larger initial library improves early coverage of useful transformation patterns.
When the library reaches a reasonable size, the diversity is already high in practice because each experience contains multiple feature combinations.
At this point, Stage II (refinement and diversity control) and Stage III (closed-loop write-back) dominate the later improvements, so the method is not overly sensitive to the exact initial library size.

\subsection{Transferability Across LLMs}

We evaluate whether our experience-evolution framework transfers across different policy LLMs.
We replace the policy LLM used for sequence generation, including API-based and open-source models.

\begin{table}[htbp]
\centering
\caption{Transferability across policy LLMs. We report downstream performance (higher is better).}
\label{tab:llm_transfer}
\resizebox{\linewidth}{!}{
\begin{tabular}{lccccc}
\toprule
\textbf{Policy LLM} & \makecell{\textbf{SpectF}} & \makecell{\textbf{Amazon}\\\textbf{Employee}} & \makecell{\textbf{Higgs}\\\textbf{Boson}} & \makecell{\textbf{SVMGuide3}} & \makecell{\textbf{OpenML}\\\textbf{586}} \\
\midrule
Llama-3.1-8B & 84.21\% & 94.16\% & 71.88\% & 83.78\% & 0.7491 \\
Llama-3.2-3B & 83.57\% & 94.28\% & 72.04\% & 83.78\% & 0.7705 \\
Llama-4       & 83.57\% & 94.67\% & 72.09\% & 87.20\% & 0.7539 \\
GPT-4o       & 87.16\% & 94.41\% & 72.29\% & 87.68\% & 0.7406 \\
o1       & 85.22\% & 94.68\% & 71.92\% & 86.71\% & 0.7434 \\
o3       & 84.21\% & 94.64\% & 71.09\% & 86.23\% & 0.7768 \\
DeepSeek-V3       & 84.50\% & 94.69\% & 72.02\% & 85.86\% & 0.7643 \\
Claude-3.7       & 87.67\% & 94.75\% & 72.59\% & 87.57\% & 0.7206 \\
Qwen3-4B     & 85.97\% & 94.69\% & 72.09\% & 88.05\% & 0.7652 \\
\bottomrule
\end{tabular}}
\end{table}

\textbf{Table~\ref{tab:llm_transfer}} shows that the framework works well across a wide range of LLMs.
The overall performance differences are not large, which suggests that the main benefit comes from the refined experience context rather than a specific model choice.
At the same time, stronger models can still provide additional gains in some datasets.
We observe that more capable LLMs tend to produce sequences that better match the dataset structure and the downstream task, leading to slightly higher performance.
This indicates that model capability and context quality are complementary: the experience library provides a strong, reusable prior, and stronger LLMs can better use the same context to generate more task-relevant sequences.

\textbf{Insight.} These results support the practicality of our approach.
Because the optimization target is the context data (the evolving experience library) instead of model parameters, the same pipeline can be applied to both closed-source and open-source LLMs with minimal changes.
This also makes the method flexible in deployment: users can select an LLM based on availability and cost~\cite{wang2025mixllm, bai2025learning}, while still benefiting from the same data-centric experience evolution.

\subsection{Robustness and Interpretability}
We evaluate robustness by fixing the transformed feature set produced by our method and varying only the downstream model.
All other settings are kept unchanged.

\begin{figure}[htbp]
    \centering
    \begin{subfigure}[t]{0.2\textwidth}
        \centering
        \includegraphics[width=\linewidth]{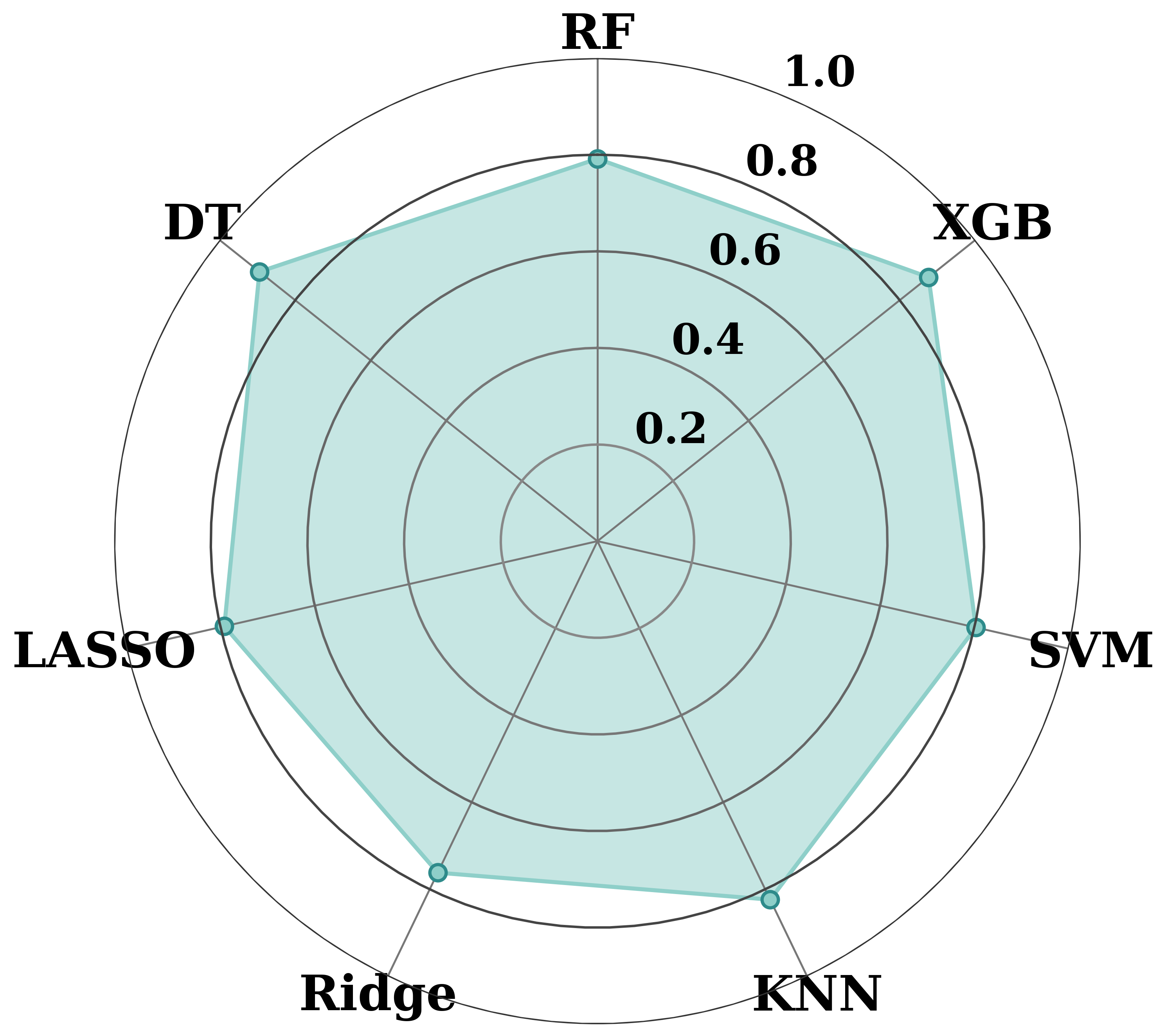}
        \caption{\textbf{SpectF}.}
        \label{fig:robust_spectf}
    \end{subfigure}
    \hspace{0.03\linewidth}
    \begin{subfigure}[t]{0.2\textwidth}
        \centering
        \includegraphics[width=\linewidth]{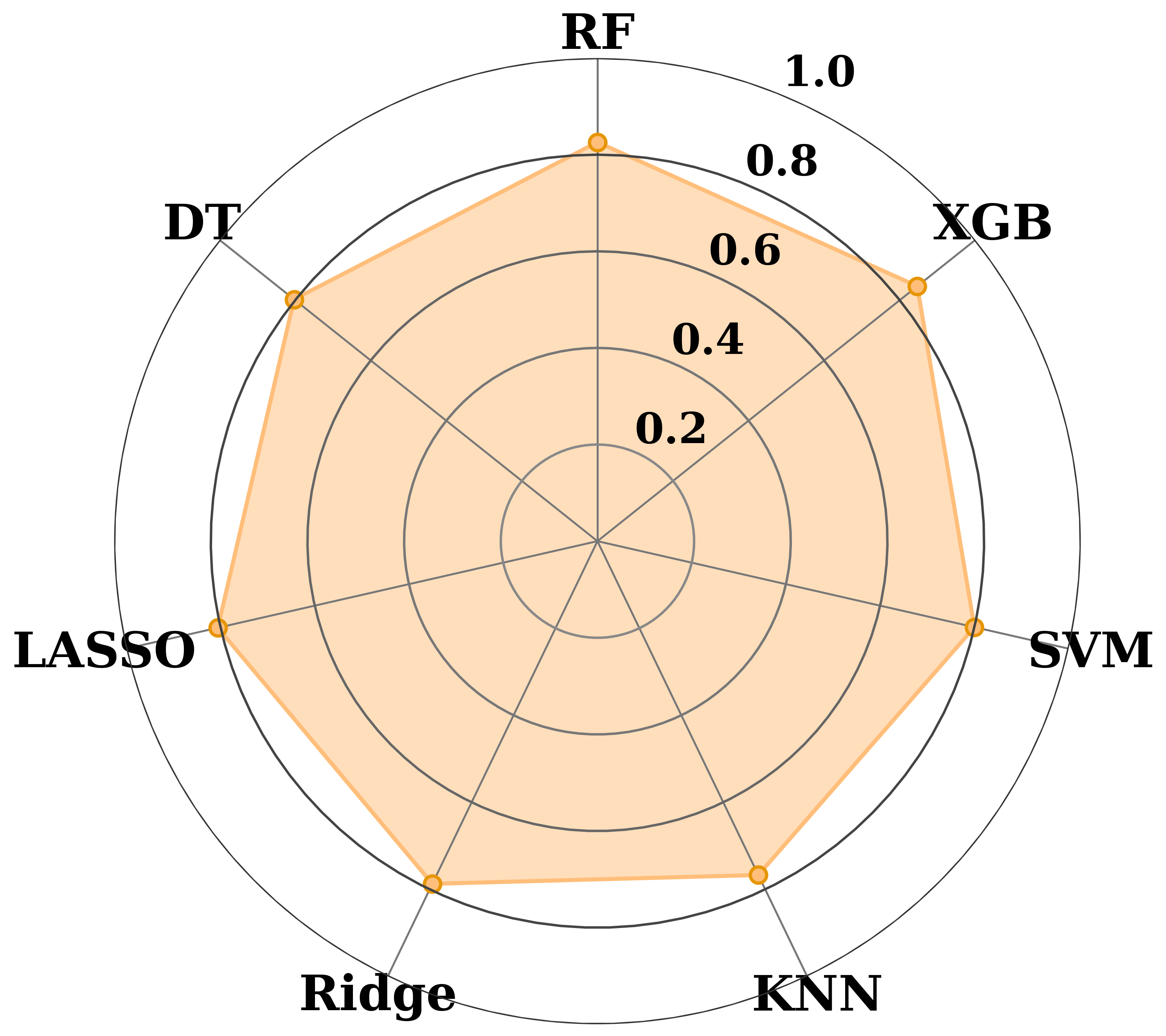}
        \caption{\textbf{SVMGuide3}.}
        \label{fig:robust_ds2}
    \end{subfigure}
    \caption{\textbf{Robustness across downstream models.} The radar plots report performance across different models.}
    \label{fig:robustness_radar}
\end{figure}

\textbf{Figure~\ref{fig:robustness_radar}} shows that the performance remains consistent across different downstream machine learning models.
Our transformed features provide stable performance and do not depend on a particular model.
This suggests that our experience-conditioned generation generally learns useful transformations rather than overfitting to one specific model.

\textbf{Insight.}
This robustness is consistent with our design.
Stage II emphasizes validation checks and redundancy control, which reduces invalid feature combinations.
The closed-loop write-back further accumulates reusable patterns that transfer across downstream models, leading to stable gains in different learning biases.

\subsection{LLM Behavior Analysis}
We found some interesting generation behaviors of LLMs in feature transformation by collecting the produced sequences and measuring their structural statistics.

\begin{figure}[htbp]
    \centering
    \begin{subfigure}[t]{0.43\linewidth}
        \centering
        \includegraphics[width=\linewidth]{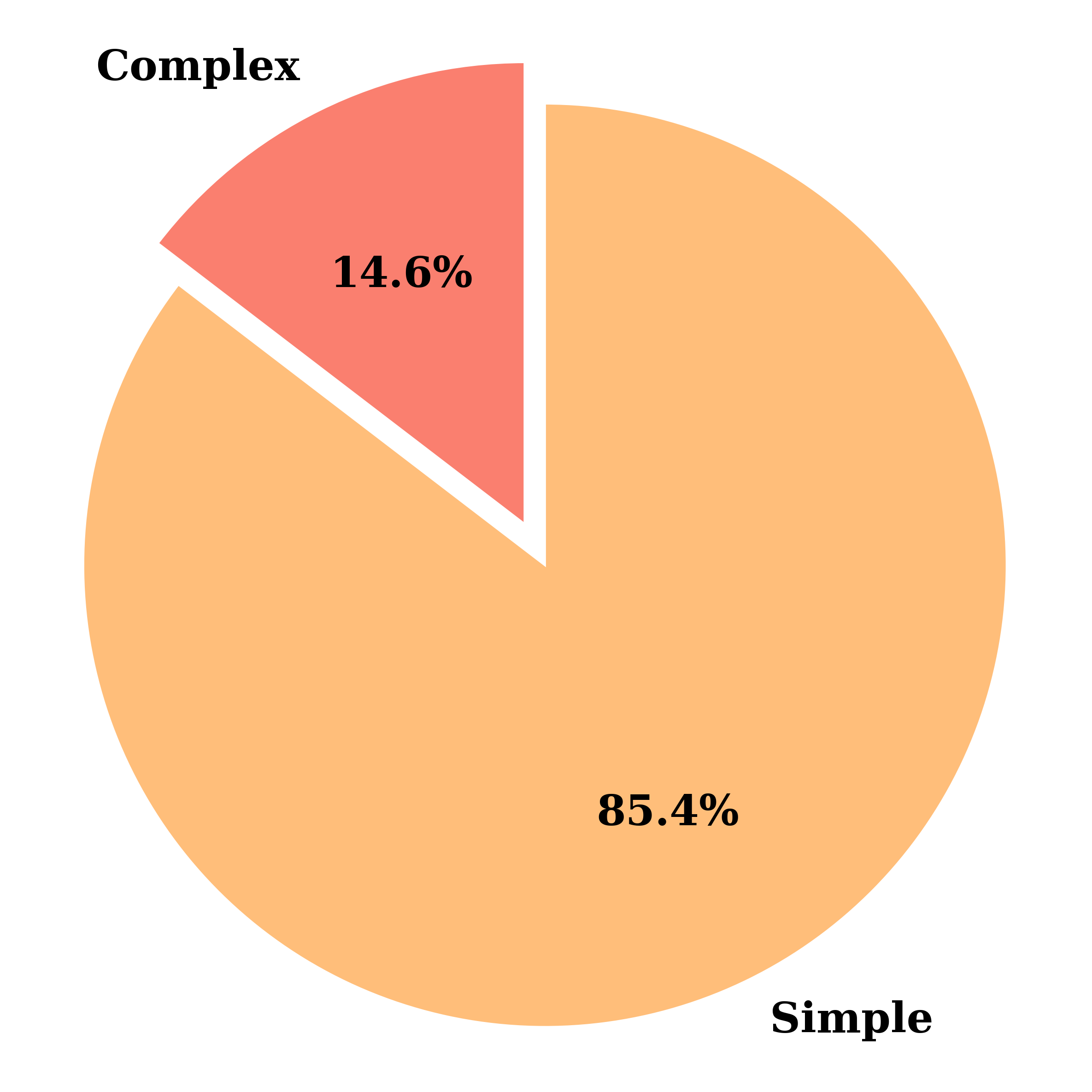}
        \caption{\textbf{LLM-generated sequences.}}
        \label{fig:op_ratio_llm}
    \end{subfigure}
    \hspace{0.03\linewidth}
    \begin{subfigure}[t]{0.44\linewidth}
        \centering
        \includegraphics[width=\linewidth]{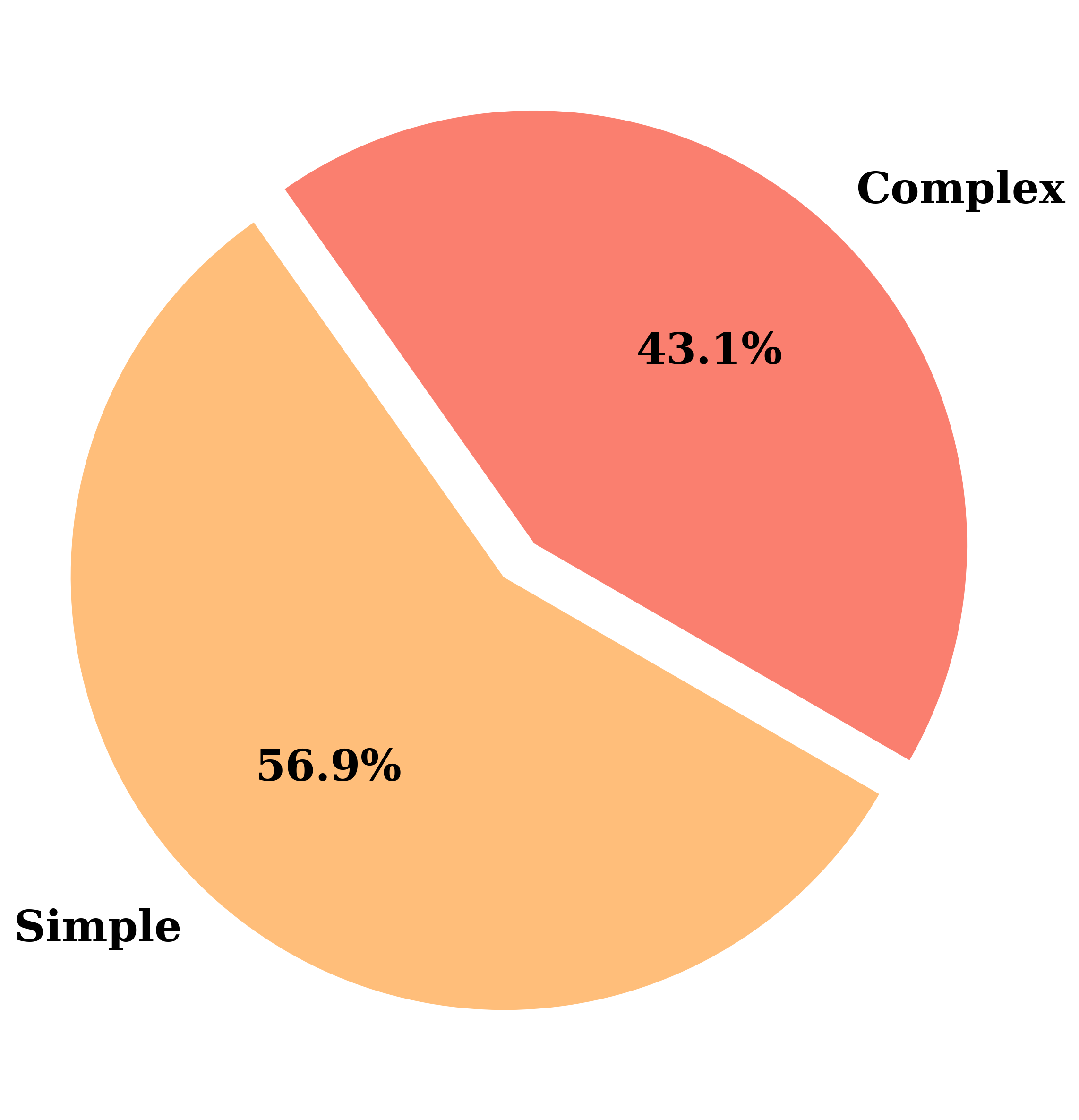}
        \caption{\textbf{RL exploration results.}}
        \label{fig:op_ratio_rl}
    \end{subfigure}
    \caption{Usage ratio of simple vs. complex operators.}
    \label{fig:operator_usage}
\end{figure}

\textbf{Finding 1: LLMs prefer simpler operators.}
We group operators into \emph{simple} ones (linear and basic scaling) and \emph{complex} ones (nonlinear and distribution-shaping transforms).
\textbf{Figure~\ref{fig:operator_usage}} compares the operator usage between the LLM-generated sequences and the RL exploration results.
LLMs use many simple operators, whereas RL exploration uses complex operators more frequently.
This suggests that LLMs tend to stay in conservative and easy-to-execute regions of the transformation space. This result shows that LLMs prioritize correct and stable answers rather than innovations.

\textbf{Finding 2: LLMs often keep original features in engineered datasets.}
In datasets that already contain engineered features, we observe that LLMs generate fewer aggressive transformations and use original features more frequently.
In \textbf{Figure~\ref{fig:feature_usage_refine}}, the LLM prefers the first 5 original features but ignores the last 20 artificial features.
This indicates that the LLM can recognize strong features from the context and implicitly performs a form of feature selection.
This behavior can be beneficial when further transformation is unnecessary or may introduce noise.

\textbf{Finding 3: apparent randomness does not indicate high effective diversity.}
Although temperature sampling leads to different surface forms, the effective diversity of generated sequences can still be low.
Results show that repeated sampling produces many near-duplicate patterns.
This motivates our entropy-guided selection and enhancement steps, which explicitly increase coverage and reduce redundancy in the experience library.

\section{Related Work}

Feature Transformation (FT) improves feature representations by applying mathematical operations to original features to boost downstream performance.
The key difficulty is the combinatorial explosion, which motivates automatic methods.

\textbf{Automated Feature Transformation}
Prior methods fall into two paradigms.
\textbf{(1) Discrete decision-making methods} treat FT as searching a discrete space where each step selects an operator and operands, and multiple steps form a trajectory.
Early systems generate many candidates and then select informative ones to control redundancy~\cite{katz2016explorekit}.
To improve search efficiency, later works adopt structured discrete strategies such as heuristic search, bandits, and other optimization techniques~\cite{kanter2015discrete1,khurana2016discrete2,tran2016discrete3}.
Evolutionary algorithms further enhance exploration by maintaining candidate populations and iterative mutation/selection~\cite{zhu2022evolutionary1}.
Reinforcement learning models FT as sequential decision-making and learns policies from downstream rewards~\cite{wang2022group,ying2023self,ying2025topology}, whereas traceable/evolutionary variants emphasize interpretability and controlled search dynamics~\cite{xiao2023traceable,xiao2024evolutionary2}.
\textbf{(2) Continuous optimizing methods} embed transformation sequences into continuous representations and optimize in latent space~\cite{wang2023reinforcement,zhu2022difer,gong2025sculpting,bai2025privacy,ying2024feature,wang2025knockoff,ying2024revolutionizing,ying2024unsupervised}.
They can smooth optimization~\cite{ying2025distribution}, but must map optimized embeddings back to valid executable programs, which may introduce a representation mismatch.

\textbf{LLMs for Feature Transformation}
LLMs have been used to generate symbolic and program-like feature constructions from dataset context or metadata~\cite{hollmann2023large, xie2025transformer}.
Related work also uses in-context prompting to produce explicit transformation sequences~\cite{gong2025unsupervised,wang2025llm}, sometimes combined with evolutionary search to maintain diversity and improve downstream performance~\cite{gong2024evolutionary}.
However, most LLM-based FT methods remain inference-time prompt-centric~\cite{gong2025agentic,fu2025autonomous, xie2026agent}: demonstrations are usually fixed or manually curated, and useful patterns are not systematically accumulated across tasks.
This often leads to instability, redundancy, or biased exploration, motivating approaches that treat demonstrations as updatable experience.

\textbf{Context-space Experience Distillation and Closed-loop Optimization}
Recent studies show that frozen LLMs can be guided by optimizing the \emph{context} as data, distilling high-quality experiences into a token-level prior that guides future generations~\cite{cai2025trainingfree}.

\section{Conclusion}


We studied LLM-driven feature transformation from a data-centric view, where the key ``data'' for guiding an LLM is the few-shot context rather than model parameters.
Across diverse tabular benchmarks, we find that closed-loop write-back is important for stability: compared with one-shot prompting, updating the experience library over iterations gives more consistent improvements under the same evaluation budget.
Ablation studies show that CoT-style organization helps compose useful transformations, while validity checking, enhancement, and diversity-aware selection reduce invalid or redundant patterns and make feature usage more balanced.
We also observe clear generation patterns that help explain LLM behaviors in feature transformation.
Finally, the framework works well with both API-based and open-source LLMs and stays robust across downstream evaluators, suggesting that optimizing context data is a practical way to improve LLM-driven feature transformation.


\bibliographystyle{ACM-Reference-Format}
\bibliography{sample-base}


\end{document}